\documentclass[11pt]{article}
\usepackage{CJKutf8}
\AtBeginDvi{\input{zhwinfonts}}
\usepackage{amsmath,textcomp}
\usepackage{times}
\usepackage{enumitem}
\usepackage{url}
\usepackage{latexsym}
\usepackage{graphicx}
\usepackage[usenames,dvipsnames,svgnames,table]{xcolor}

\newcommand{\boxfigure}[1]%
        {\framebox[\textwidth]{%
         \parbox{0.95\textwidth}{#1}}}

\newtheorem%
     {theorem}{Theorem}[section]
\newtheorem%
     {corollary}[theorem]{Corollary}
\newtheorem%
     {proposition}[theorem]{Proposition} 
\newtheorem%
     {lemma}[theorem]{Lemma} 

\newtheorem%
     {exerciseAux}[theorem]{Exercise}
\newenvironment%
     {exercise}{\begin{exerciseAux}\rm}{\end{exerciseAux}}

\newtheorem%
     {beispielAux}[theorem]{Beispiel}
\newenvironment%
     {beispiel}{\begin{beispielAux}\rm}{\end{beispielAux}}

\newtheorem%
     {definitionAux}[theorem]{Definition}
\newenvironment%
     {definition}{\begin{definitionAux}\rm}{\end{definitionAux}}
\newtheorem%
     {remarkAux}[theorem]{Remark}
\newenvironment%
     {remark}{\begin{remarkAux}\rm}{\end{remarkAux}}

\newtheorem%
     {exampleAux}[theorem]{Example} 
\newenvironment%
   {example}{\begin{exampleAux}\parskip=0in\rm}{\parskip=0.1in\end{exampleAux}}

\newtheorem%
     {examplesAux}[theorem]{Examples} 
\newenvironment%
     {examples}{\begin{examplesAux}\rm}{\end{examplesAux}}

\newtheorem%
     {constructionAux}[theorem]{Construction} 
\newenvironment%
     {construction}{\begin{constructionAux}\rm}{\end{constructionAux}}

\def\qed{\hfill{\boxit{}}
  \ifdim\lastskip<\medskipamount \removelastskip\penalty55\medskip\fi}
\def\proofend{\hfill{\boxit{}}
  \ifdim\lastskip<\medskipamount \removelastskip\penalty55\medskip\fi}
\long\def\boxit#1{\vbox{\hrule\hbox{\vrule\kern3pt
                  \vbox{\kern3pt#1\kern3pt}\kern3pt\vrule}\hrule}}

\newenvironment%
     {genass}%
        {\medbreak\noindent{\bf General Assumption.\enspace}\it}%
        {\ifdim\lastskip<\medskipamount \removelastskip\penalty55\medskip\fi}

\newenvironment
     {comeqns}%
        {\vspace{-.5ex}      
         \[   %
              \begin{array}{lcl@{\qquad}rl}}%
        {\end{array}
         \]
         \vspace{-.5ex}}

\newenvironment
     {dbenum}%
        {\vspace{-.5ex}      
         \[   %
              \begin{array}{r@{\quad}l@{\qquad\qquad}r@{\quad}l}}%
        {\end{array}
         \]
         \vspace{-.5ex}}

\newcounter{zaehler}

\newenvironment
     {paireqns}%
        {\vspace{-.5ex}      
         \[   %
              \begin{array}{rcl@{\qquad}rcl}}%
        {\end{array}
         \]
         \vspace{-.5ex}}

\newcommand{\pref}{\mbox{\sl pref\/}}
\newcommand{\suff}{\mbox{\sl suff\/}}
\newcommand{\INF}{\mbox{\sl inf\/}}
\newcommand{\SPref}{\mbox{\sl SPref\/}}
\newcommand{\SSuff}{\mbox{\sl SSuff\/}}
\newcommand{\parent}{\mbox{\sl msc\/}}
\newcommand{\Inf}{\mbox{\sl Inf\/}}

\usepackage{authblk}
\usepackage{hyperref}
\title{Corpus analysis without prior linguistic knowledge - unsupervised mining of phrases and subphrase structure}

\author[1,2]{Stefan Gerdjikov {\thanks{st\_gerdjikov@abv.bg}}} 
\author[1]{Klaus U. Schulz {\thanks{schulz@cis.uni-muenchen.de}}}
\affil[1]{CIS, Ludwig Maximilians University, Munich, Germany }
\affil[2]{FMI, University of Sofia "St. Kliment Ohridski", Sofia, Bulgaria }

\begin{document}

\maketitle
\begin{abstract}
When looking at the structure of natural language, ``phrases'' and ''words'' are central notions. 
We consider the problem of identifying such ``meaningful subparts'' of language of any length and underlying composition principles in a completely corpus-based and language-independent way without using any kind of prior linguistic knowledge. 
Unsupervised methods for identifying ``phrases'', mining subphrase structure and finding words in a fully automated way are described. This can be considered as a step towards automatically computing a ``general dictionary and grammar of the corpus''. We hope that in the long run variants of our approach turn out to be useful for other kind of sequence data as well, such as, e.g., speech, genom sequences, or music annotation. Even if we are not primarily interested in immediate applications, results obtained for a variety of languages show that our methods are interesting for many practical tasks in text mining, terminology extraction and lexicography, search engine technology, 
and related fields. 
\end{abstract}


\parskip 1mm

\section{Introduction}

While there is no complete agreement on what the structural subparts of natural language exactly are, there is a general understanding that 
natural language is built in a compositional way using characteristic classes of substrings (or utterances) such as phrases, words, and morphems. In natural language processing (NLP) and related fields, all these units play an important practical role.  
Phrases are used, e.g., in Statistical Machine Translation (SMT)~\cite{ON04,PK10}, Text Classification~\cite{S01,DIK11}, and 
Information Retrieval~\cite{MRS07,LY07,ZLC13}. Words (``tokens'' and ``types'') represent the basis for text indexing approaches in Information Retrieval, and since centuries linguists collect words in general and specific lexica, including typical contexts and collocations. Morphems carry information on part-of-speech, syntax and semantic roles ~\cite{LY07}. 

In this paper we present results from an ongoing research project which is centered around the problem of {\em how to detect structural subparts of natural (or other) language} and is characterized by the following {\em methodological principles}.
\begin{enumerate}
\item Try to detect {\em meaningful} units of {\em any length}, as opposed to, e.g., fixed length $n$-grams of words or letters. 
\item Only use {\em corpora (or sequence data) as an empirical basis} to find these units.
\item When analyzing corpora, do {\em not use any kind of prior linguistic knowledge} and {\em no knowledge on the nature/functionality of a particular symbol} (such as blank, hyphen, brackets, uppercase vs. lowercase etc.) but rather simple mathematical assumptions.
\item Proceed in a completely {\em unsupervised} way.
\end{enumerate}
While we acknowledge the success of NLP approaches based on $n$-gram statistics, it is clear that using units of fixed length does not naturally lead to meaningful subparts. In anatomic terms, our aim here is to find the ``muscles, organs, and nerves'' of a given language corpus. 
The other above principles are related in the sense that they stress purely empirical evidence. Corpora, from our point, are just raw sequences of symbols. For analyzing corpora we do not use prior linguistic knowledge since we want to see what ``the corpus structure itself tells us about language''. This might help to compare existing linguistic knowledge and hypotheses with ``purely empirical'' corpus-driven evidence. Furthermore, principles and methods obtained from this puristic approach can be applied to {\em any} (not necessarily natural) language and might contribute to the analysis of other kind of sequence data such as biological sequences, sequence of gestures, animal language, music annotation, and others. We avoid any kind of supervision since we are not primarily interested in the path from prior human linguistic knowledge to the analysis of data, but rather on the question how structure can be detected and then ``destilled'' to knowledge. The focus of this paper is the mining of phrases, subphrase structure, and words. From a practical side, one vision is to automatically and ``online'' create a kind of ``general dictionary and grammar of a given corpus'' that collects, connects and relates all interesting language units of the corpus, with a direct reference to all occurrences. In the Conclusion we add a remark on possible interaction of our approach with unsupervised morphology induction approaches, such as ~\cite{Goldsmith01,Z08,creutz2005unsupervised,soricut2015unsupervised}.

As a matter of fact, this programmatic approach is by no means new. As an early influential advocate, Zellig Harris \cite{Harris54,Harris91} has stressed that 
corpus analysis based on mathematical principles should be used to derive empirical evidence for linguistic hypotheses. Later, Maurice Gross \cite{Gross1993,Gross1999a}, Franz Guenthner \cite{Guenthner2005}, Johannes Goller \cite{Goller2010} and others followed this line. 
Corpus analysis is often used to find lexico-semantic relations \cite{storjohann2010lexical}. Further related work is described in Section~\ref{Sec5}. 
One algorithmic contribution of our work is the use of a special 
index structure for corpora as a basis: symmetric directed acyclic word graphs (SDAWGs) \cite{Inenaga01} represent a generalization of suffix trees \cite{Ukkonen95}. Using the SDAWG index of a corpus, for any substring of the corpus we have immediate access to all {\em left and right contexts of arbitrary length}. In this way we see ``how'' a substring occurs in the corpus (i.e., in which contexts and larger sequences, how often), which provides an excellent basis for recognizing meaningful units and connected tasks.

When asking for the composition of natural language in terms of meaningful subparts, a fundamental question is how small parts are combined to form larger units.  
Most grammar approaches (e.g., context-free grammars) are ``concatenation based'' in the sense that always {\em disjoint, non-overlapping} parts are glued together to build larger units. In contrast, our base assumption is that in general ``phrases'' can overlap, and combination principles go beyond plain concatenation.

Our base view of language composition is briefly explained in Section~\ref{Sec1}. 
Using this view, Section~\ref{Sec2} introduces mathematical principles and an algorithm for recognizing meaningful units and decomposing sentences into ``phrases''. 
We illustrate the sentence decompositions and ``phrases'' obtained for a variety of distinct languages (English, German, French, Spanish, Dutch, Chinese) and corpora (including Wikipedia, Europarl, Medline). 
Section~\ref{Sec3a} introduces principles for further decomposing ``phrases''. The base units found in this way (in most cases) are ``words''. Recall that we start from a point where we do not have the concept of a ``word''. Again we presents results obtained for distinct corpora and languages. 
In Section~\ref{Sec4} we discuss some possible applications of the phrase detection mechanisms: first we show how subphrase structure can be used to find a ranked spectrum of ``most characteristic words'' of the given corpus, and to extract paradigmatic word networks based on co-occurrence of terms in phrases. We also look at stylistic corpus analysis, terminology extraction, and automated query expansion for search engines.
In Section~\ref{Sec5} we briefly comment on related work. Section~\ref{Sec6} gives a conclusion and comments on future work.

\section{Phrase model - intuitions and motivation}\label{Sec1}

As a starting point we look at the notion of a ``phrase''. When we want to recognize ``phrases'' in a corpus-driven way we need as a starting point some primitive base assumptions how ``meaningful phrases'' can be recognized and distinguished from arbitrary, ``meaningless subsegments'' of language. 
In our view, recognition of phrases should proceed in parallel with the recognition of those language units that connect and interlink phrases. Our main motivation comes from psycholinguistics ~\cite{HW03,SWC06,T07}: 
 \emph{Young children recognise function words but do not use them in the beginning.}
This suggests that there are some natural principles 
that govern the structure of natural language that are strongly related to the use of function words. Children are able to detect these regularities, and to distinguish between language parts that serve to {\em combine and connect} meaningful parts - function words - on the one-hand side, and {\em meaningful} parts - content words and phrases - on the other side.

Function words, or closed-class words, alone bear little sense,~\cite{F52}. Their
main function is to bind other words or expressions in a grammatically correct way 
to form more complex meaningful language units. To suit these needs of the language, only a few
function words, e.g. prepositions, articles, suffice. Thus, although some specific word may act as a function word in a special domain, generally the set of function words is stable in every particular language.

In our situation we do not assume that we know the set of function words, and the notion of a ``word'' is partially misleading: we do not assume that ``phrase connectives'' are always words, we even do not have 
a notion of ``word'' at the beginning. Yet, for convenience, we say ``function word'' when we mean ``phrase connective''. 
We assume that the corpus consists of sentences\footnote{Other small meaningful text units like paragraphs also conform with our model.} which make sense. The remaining assumptions we make about phrases and function words in the corpus are:   
\begin{enumerate}[label=(A\arabic*),start=1]
\item\label{lab_amb} Phrases occur in different contexts.
\item\label{lab_dec} Sentences are built of (overlapping) phrases.
\item\label{lab_str} Phrases overlap on function words.
\end{enumerate}
The first assumption means that the corpus provides structural evidence that helps to recognize phrases.
The circumstance that the corpus consists of sentences that make sense is important for the second and third 
assumptions\footnote{However, this is not a prerequisite for the mathematical approach described in the next section.}. 
Assumption~\ref{lab_dec} follows the intuition that the sense of a sentence is determined by the sense of the individual phrases and the way they are composed,
whereas the final assumption reflects the fact that function words bind smaller meaningful language units to build larger ones.
In our case, the smaller meaningful language units are the phrases we are looking for, and the larger ones are
the sentences in the corpus.

Since function words alone, to a large extent, are lacking sense, we can fairly approximate the sense
of the sentence with the sense accumulated by the individual phrases. Thus, we do not need to bother
of modelling what \emph{sense} is, rather we will be concerned with the issue how regular w.r.t. the given corpus
the bindings between the phrases are. We try to summarise this intuition in the following two principles:
\begin{enumerate}[label=(P\arabic*),start=1]
\item\label{P1} The function words occur regularly as the boundaries of phrases in the corpus. 
\item\label{P2} A particular sentence should exhibit these regularities as good as its structure w.r.t. the corpus allows.
\end{enumerate}

\section{Computing phrases and function words}\label{Sec2}

Following the considerations of the previous section, we first look at the problem of how to compute the phrases and function words of a corpus. In the first subsection we develop an algorithm to solve this task. Afterwards we present the results obtained for distinct languages and corpora.  

\subsection{Algorithmic principles}\label{Sec2Sub1}

We consider a corpus, $\mathcal{C}$, consisting of a finite number of sentences. The individual sentences are merely
nonempty strings $S_n\in \Sigma^+$ over a finite alphabet $\Sigma$. In practice, the alphabet $\Sigma$ is the set of all the symbols, i.e. letters,
punctuation, white space, etc., that were used to create the corpus. 
According to Assumption~\ref{lab_amb}, the phrases are among those \emph{strings} $S\in\Sigma^*$ that occur in different contexts within the corpus $\mathcal{C}$. Formally, for a phrase $S$ the following two properties must hold:
\begin{enumerate}
\item\label{m_left} there are distinct $a,b\in \Sigma$ such that $aS$ and $bS$ occur in $\mathcal{C}$, or $S$ is a prefix of some of the sentences $S_n$, and
\item\label{m_right} there are distinct $c,d\in \Sigma$ such that $Sc$ and $Sd$ occur in $\mathcal{C}$, or $S$ is an suffix of some of the sentences $S_n$.
\end{enumerate}
Strings with the above two properties are called {\bf general phrase candidates} of the corpus, 
the set of general phrase candidates is denoted $\mathcal{G}(\mathcal{C})$. For example, if the strings \texttt{unfortunate\_} and \texttt{\_fortunately} occur as part of some sentences in the corpus\footnote{We use the symbol ``$\_$'' to make a white space explicit.}, then they witness that \texttt{fortunate} can be followed by at least two distinct letters, \texttt{l} and \texttt{\_}, and similarly it can be preceded by at least two distinct letters, \texttt{n} and \texttt{\_}. Therefore \texttt{fortunate} is a general phrase candidate in this case. However, if it turned out that the string \texttt{ortuna} is always preceded by the letter \texttt{f} in the corpus, then the string \texttt{ortuna} will be \emph{not} a member of $\mathcal{G}(\mathcal{C})$. The same would be true if 
\texttt{ortuna} was always followed by a \texttt{t} in the corpus.

According to \ref{lab_dec}, a sentence $S_n$ in the corpus can be decomposed into a sequence of nonempty strings $(P_1,P_2,\dots,P_{k_n})$  with $k_n>1$ such that $P_i\in \mathcal{G}(\mathcal{C})$. For instance, if a sentence starts like 
\begin{quotation}\texttt{We\_are\_not\_making\_the\_connection\_...} 
\end{quotation} 
it may be the case that $P_1=\texttt{We\_are\_not\_}$, $P_2=\texttt{\_are\_not\_making\_the\_}$, $P_3=\texttt{\_the\_connection}$. In this case $P_1$ and $P_2$ overlap on the string $F_1=\texttt{\_are\_not\_}$, whereas $P_2$ and $P_3$ overlap on the string $F_2=\texttt{\_the\_}$. In general, $P_i$ and $P_{i+1}$ overlap on a common possibly empty string $F_i\in \Sigma^*$, i.e. $P_i=P_i'\circ F_i$ and $P_{i+1}=F_i\circ P_{i+1}''$, such that:
\begin{equation}
S_n=P_1\circ F_1^{-1}\circ P_2\circ F_2^{-1}\dots\circ F_{k_n-1}^{-1}\circ P_{k_n}.\nonumber
\end{equation}
We call the $F_i$'s {\bf function strings}. Upon termination of our approach will declare some of these strings
to be \emph{function words}. Recall that if $P_i, P_{i+1}$ are in the set of general phrase candidates $\mathcal{G}(\mathcal{C})$, then
each suffix of $P_i$ satisfies Property~\ref{m_right} and each prefix of $P_{i+1}$ satisfies Property~\ref{m_left}.
In particular, since $F_i$ is a suffix of $P_i$ and $F_i$ is a prefix of $P_{i+1}$, it follows that each function string $F_i$ is always a general phrase candidate.

In general, there can be many decompositions, $(P_1,P_2,\dots,P_{k_n})$, for a given sentence $S_n$. According to Principle~\ref{P2} 
we should select the one(s) that exhibit(s) the most regular sequence of function strings, $(F_1,F_2,\dots,F_{k_{n}-1})$. Further, according
to Principle~\ref{P1}, the function strings should occur regularly as the boundaries of phrases. We will incorporate these two principles
in terms of appropriate probability and likelihood measures. 

To do so, let us assume for a moment that we have already determined a multiset of phrases $\mathcal{P}\subseteq\mathcal{G}(\mathcal{C})$. Based on~\ref{P1} we are going to assess the property of a function string $F$ to be a function word in the corpus. Next, based on~\ref{P2}, we are going to determine the best decomposition of each sentence in the corpus as a sequence of general phrase candidates. This second step will allow us to refine (redefine) the multiset of phrases we have started with. We can then iterate this process until the multiset of phrases stabilizes. 
Based on the obtained probabilities and the eventual sentence decompositions we define the ultimate set of function words and phrases.

Formally, to assess the property of $F$ to be a function string w.r.t. the current multiset of phrases $\mathcal{P}$, we simply consider how often $F$ occurs as a proper prefix, suffix or
proper infix in $\mathcal{P}$:
\begin{eqnarray}
\pref(F|\mathcal{P}) = |\{ F\circ S\in\mathcal{P}\,|\, S\in\Sigma^+\}|\nonumber\\
\suff(F|\mathcal{P}) = |\{ S\circ F\in\mathcal{P}\,|\, S\in\Sigma^+\}|\nonumber 
\end{eqnarray}
\begin{equation}
\INF(F|\mathcal{P}) = |\{ S_1FS_2\in\mathcal{P}\setminus\{F\} \,|\, (S_1, S_2)\in(\Sigma^*)^2\}| \nonumber.
\end{equation}
This statistics accounts for multiplicities. For example, if $F=\texttt{\_the\_}$ and the phrase $P=\texttt{\_the\_connection\_}$ occurs $12$ times in $\mathcal{P}$, then $P$ will contribute with $12$ units to $\pref(\texttt{\_the\_}|\mathcal{P})$. Now, we define the probability of a string $F\in\mathcal{G}(\mathcal{C})$ to be a function word w.r.t. to a given multiset $\mathcal{P}\subseteq\mathcal{G}(\mathcal{C})$ as the empirical probability that $F$ is both a proper prefix and a proper suffix of a string in $\mathcal{P}$:
\begin{equation}\label{eq:definition}
p_{fw}(F|\mathcal{P}) =p_{pref}(F|\mathcal{P})\cdot p_{suf}(F|\mathcal{P})
\end{equation}
where
\begin{equation}
 p_{pref}(F|\mathcal{P})=\frac{\pref(F|\mathcal{P})}{\INF(F|\mathcal{P})},\, \quad p_{suf}(F|\mathcal{P})=\frac{\suff(F|\mathcal{P})}{\INF(F|\mathcal{P})}.\nonumber
\end{equation} 
To model \ref{P2} we optimize the likelihood w.r.t. $p_{fw}(.|\mathcal{P})$. This means, that given a sentence $S_n\in\mathcal{C}$ we search for those sequences of strings $(P_1,P_2,\dots ,P_{k_n})$ ($k_n>1$) whose sequence of overlaps $\overline{F}=(F_1,F_2,\dots,F_{k_{n}-1})$ maximises the function:
\begin{equation}\label{eq:likelihood}
\ell(\overline{F}|S_n,\mathcal{P},\mathcal{C})=\prod_{i=1}^{k_n-1}p_{fw}(F_i|\mathcal{P}).
\end{equation}
Determining the maxima of Equation~\ref{eq:likelihood} supplies us with (an) optimal decomposition(s) $(P_1,P_2,\dots,P_{k_n})$ for each sentence $S_n$ in the corpus ${\cal C}$.
We call a substring $S_n[i..j]$ of $S_n$ an \emph{optimal phrase candidate for $S_n$} if it belongs to some optimal decomposition for $S_n$. We call the overlap of two consecutive strings in an optimal decomposition for $S_n$ an \emph{optimal overlap}.
Now, we are in a position to reconsider the multiset of phrases $\mathcal{P}$ we have started with. The intuition is, that if we have made a reasonable even if not perfect 
guess for the set of phrases in the first step, then we should have obtained good scores for the typical function words. Thus, in the second step, optimising the likelihood across the sentences the function words will guide us to optimal strings which better describe the proper phrases in $\mathcal{C}$.
Following these considerations we define the new multisets
\begin{eqnarray}
\mathcal{F}(\mathcal{P}) &:=&\{S_n[j_1..j_2]\,|\, S_n[j_1..j_2] \text{ optimal overlap in a sentence } S_n\} \nonumber\\
\mathcal{P}' &:=& \{S_n[j_1..j_2]\,|\, S_n[j_1..j_2] \text{ optimal phrase candidate for a sentence } S_n\}.\nonumber 
\end{eqnarray}
Although the approach described above reflects  principles \ref{lab_amb} - \ref{lab_str} as well as \ref{P1} and~\ref{P2}, there is one subtle point that it overlooks. 
The optimisation of the likelihood implicitly struggles for shorter sequences of phrases. For instance, if 
\begin{eqnarray*}
P_1 &=& \texttt{\_are\_not\_making\_the\_}\\
P_2 &=& \texttt{\_the\_connections}\\
P &=& \texttt{\_are\_not\_making\_the\_connections}
\end{eqnarray*}
are three general phrase candidates, the optimisation of Equation~\ref{eq:likelihood} in situations like \texttt{We\_are\_not\_making\_the\_connections...} will always prefer the longer phrase, $P$, rather than the two shorter overlapping phrases $P_1$ and $P_2$. Thus, $P=\texttt{\_are\_not\_making\_the\_connections}$ contributes to the count $\INF(\texttt{\_the\_}|\mathcal{P})$, but it will not contribute to the counts  $\pref(\texttt{\_the\_}|\mathcal{P})$ and $\suff(\texttt{\_the\_}|\mathcal{P})$. This is a problem, because if $\texttt{\_are\_not\_making\_the\_}$ and $\texttt{\_the\_connections}$ are indeed phrases, then the presence of the phrase 
$$
\texttt{\_are\_not\_making\_the\_connections}
$$ 
witnesses that $\texttt{\_the\_}$ acts as binding element in this case. However,  this example will contribute to \emph{decreasing} instead of increasing the probability $p_{fw}(\texttt{\_the\_}|\mathcal{P})$.      

To remedy this inconsistency in the model, we take additional care for the cases described above and reflect them in the probability measure for a function word. Specifically, we define the set of stable prefix (suffix) phrases of a phrase: let $P\in\mathcal{P}$. A prefix (suffix) $P'$ of $P$ is a {\em stable prefix (suffix) phrase of $P$} w.r.t. the the set of phrases ${\cal P}$ iff two conditions are satisfied:
\begin{enumerate}
\item $P'$ is a phrase in ${\cal P}$,
\item every prefix (suffix) $P''$ of $P$ of length $\vert P''\vert > \vert P'\vert$ which is a general phrase candidate is again a phrase in ${\cal P}$. 
\end{enumerate}
By $\SPref(P|\mathcal{P})$ and $\SSuff(P|\mathcal{P})$ we respectively denote the set of all stable prefix (suffix) phrases of phrase $P$. 
Now, we account for the inconsistency of the model described above in the following way. We consider phrases $P\in\mathcal{P}$ which can be
represented as two overlapping phrases $P_1,P_2\in\mathcal{P}$ but at least one of them should be stable. This means that either $P_1$ is a stable prefix phrase of $P$, or $P_2$ is a stable suffix phrase of $P$. 
The {\em overlap value} $ov(F|\mathcal{P})$ of a function string $F$ w.r.t. the set of phrases ${\cal P}$ is defined as the number of phrases, $P$, in the from $P_1\circ F^{-1}\circ P_2\in\mathcal{P}$ where $(P_1,P_2)\in\mathcal{P}^2$ such that $P_1\in \SPref(P)$ or $P_2\in \SSuff(P)$. As before we account for possible multiplicities $P$ in the set $\mathcal{P}$.
The overlap value $ov(F|\mathcal{P})$ is used to boost the empirical probabilities $p_{pref}(F|\mathcal{P})$ and $p_{suf}(F|\mathcal{P})$:
\begin{eqnarray*}
\tilde{p}_{pref}(F|\mathcal{P}) &=& p_{pref}(F|\mathcal{P})+\eta_{pref}(F|\mathcal{P})\frac{ov(F|\mathcal{P})}{\INF(F|\mathcal{P})}\\
\tilde{p}_{suf}(F|\mathcal{P}) &=& p_{suf}(F|\mathcal{P})+\eta_{suf}(F|\mathcal{P})\frac{ov(F|\mathcal{P})}{\INF(F|\mathcal{P})}
\end{eqnarray*} 
where $\eta_{i}(F|\mathcal{P}) = p_{i}(F|\mathcal{P})/(p_{pref}(F|\mathcal{P})+p_{suf}(F|\mathcal{P}))$ for $i\in \{\mbox{\sl pref,suf}\}$.
Essentially, the above formulae assume that the counts $\pref(F\vert {\cal P})$ and $\suff(F\vert {\cal P})$ are representative and thus the ratio $\pref(F\vert {\cal P})/\suff(F\vert {\cal P})$ reliably represents the true ratio of prefix-suffix property. Yet, due to cases accounted for in $ov(F|\mathcal{P})$, they are underestimated with respect to the total number of occurrences of $F$ as an infix. This is why we distribute the amount $ov(F|\mathcal{P})$ as to maximise the entropy w.r.t. the empirical $\pref(F\vert {\cal P})/\suff(F\vert {\cal P})$.
In the final model instead of $p_{pref}$ and $p_{suf}$ in Equation~\ref{eq:definition} we use $\tilde{p}_{pref}$ and $\tilde{p}_{suf}$, respectively. 

{\bf Algorithm for detecting phrases and function words}. Our complete algorithm for detecting phrases and function words proceeds in the following way. 

\noindent{\em 1. Initialisation.\ } 
We start with the multiset of phrases $\mathcal{P}_0 := \mathcal{G}(\mathcal{C})$. The multiplicity of each string is simply the total number of occurrences of the string in the corpus.

\noindent{\em 2. Iteration.\ } Afterwards we follow the spirit of the Maximum A Posteriori Principle,~\cite{TBS87,GJ08}, and compute $\tilde{p}_{pref}^{(i)}=\tilde{p}_{pref}(.|\mathcal{P}_i)$ and $\tilde{p}_{suf}^{(i)}=\tilde{p}_{suf}(.|\mathcal{P}_i)$ and define the next set of phrases as $\mathcal{P}_{i+1}=\mathcal{P}_i'$ (see above). 

\noindent{\em 3. Termination.\ }
As in other similar approaches the final goal is to obtain convergence. In our case we want to arrive at a state where the parameters $\mathcal{P}_i$, $\tilde{p}_{pref}$ and $\tilde{p}_{suf}$ stabilize. Since the probability functions $\tilde{p}_{pref}$ and $\tilde{p}_{suf}$ are entirely determined by the current set of phrases $\mathcal{P}_i$ it is enough to define a convergence criterium for the multisets $\mathcal{P}_i$. 

In order to assess the similarity between the multisets $\mathcal{P}_i$ and $\mathcal{P}_{i+1}$, we use a pure set theoretical measure. Let $m_i(P)$ denote the multiplicity of $P$ in the set $\mathcal{P}_i$, i.e. the number of times the general phrase candidate $P$ is assigned to $\mathcal{P}_i$. We can express the size of the symmetric difference $\mathcal{P}_i\Delta \mathcal{P}_{i+1}$ of the two multisets $\mathcal{P}_i$ and $\mathcal{P}_{i+1}$ as
\begin{equation*}
\|\mathcal{P}_i\Delta \mathcal{P}_{i+1}\|=\sum_P |m_i(P)-m_{i+1}(P)|.
\end{equation*}
Similarly we compute the size of the union $\mathcal{P}_i\cup \mathcal{P}_{i+1}$ of the two multisets as
\begin{equation*}
\|\mathcal{P}_i\cup \mathcal{P}_{i+1}\|=\sum_P \max(m_i(P),m_{i+1}(P)).
\end{equation*}
As a similarity measure between the multisets $\mathcal{P}_i$ and $\mathcal{P}_{i+1}$ we use the relative size
\begin{equation*}
\rho_i=\rho(\mathcal{P}_i,\mathcal{P}_{i+1})=\frac{\|\mathcal{P}_i\Delta \mathcal{P}_{i+1}\|}{\|\mathcal{P}_i\cup \mathcal{P}_{i+1}\|}.
\end{equation*}
We iterate Step~{\em 2} until:
\begin{equation}[Halting\  Criterion]
\quad \rho_i \leq \rho_{i+1} +\theta, \text{ where } \theta=10^{-6}.
\end{equation}
We will comment on this halting criterion at the end of this section.

\noindent{\em 4. Final model.\ }
Assume that the algorithm terminates after $n+1$ iterations of Step {\em 2} when for the first time $\rho_{n+1}\ge \rho_n-\theta$. 
We define: 
\begin{eqnarray*}
\mathcal{P} &=& \mathcal{P}_n\\
\tilde{p}_{pref} &=& \tilde{p}_{pref}^{(n)}\\
\tilde{p}_{suf} &=& \tilde{p}_{suf}^{(n)}\\
p_{fw}(F\mid {\cal P}) &=& \tilde{p}_{pref}(F\vert {\cal P})\cdot \tilde{p}_{suf}(F\vert {\cal P}).
\end{eqnarray*}
We define the  set of {\bf function words of the corpus}, $\mathcal{F}(\mathcal{C})$,  as the set of strings 
$F\in\mathcal{F}_n$ that satisfy the following properties: 
\begin{itemize}
\item $F$  has multiplicity at least $|\mathcal{C}|/10^3$ in $\mathcal{F}(\mathcal{P})=\mathcal{F}(\mathcal{P}_n)$,
\item 
$\tilde{p}_{pref}(F\vert {\cal P})+\tilde{p}_{suf}(F\vert {\cal P})>0.4$ and 
\item 
$\tilde{p}_{pref}(F\vert {\cal P})/\tilde{p}_{suf}(F\vert {\cal P})\in(1/4;4)$. 
\end{itemize}
The first condition expresses that $F$ not only occurs often in the corpus, but also serves as an overlap in
many different cases. The second and third conditions\footnote{Note that the inequality $\tilde{p}_{pref}(F\vert {\cal P})+\tilde{p}_{suf}(F\vert {\cal P})\le 1$ is always fulfilled.} capture the intuition that $F$ is both a typical prefix and suffix 
boundary of a phrase, and not only, say, a typical prefix.

Our experiments have shown that the particular choice of these parameters is not significant. Particularly, among the strings satisfying the above conditions almost all achieve $\tilde{p}_{pref}(F\vert {\cal P})+\tilde{p}_{suf}(F\vert {\cal P})>0.5$ whereas the ratio between the prefix and suffix property varies in spans $(1/3;3)$. Additionally, typically the strings meeting these constraints have total probability at rates:
\begin{equation*}
\tilde{p}_{pref}(F\vert {\cal P})\tilde{p}_{suf}(F\vert {\cal P})>9/100
\end{equation*} 
where the possible maximum is $25/100$ (attained for $\tilde{p}_{pref}(F\vert {\cal P})=\tilde{p}_{suf}(F\vert {\cal P})=0.5$). At the same the strings which are pruned have probabilities of one magnitude less.

We define the set of {\bf phrases of the corpus}, $\mathcal{P}(\mathcal{C})$ as those $P\in \mathcal{P}$ that (i) start with some element of $\mathcal{F}(\mathcal{C})$ or are the beginning of a sentence and at the same time (ii) end with some element of $\mathcal{F}(\mathcal{C})$ or are the end of a sentence. 

{\bf Remarks on the Halting Criterion.\ }
Naively, one could expect that the similarity $\rho_i$ (normalized size of symmetric difference) between two consecutive sets of phrases   
converges to $0$. 
However, in practice this does not happen The reason for this phenomenon are rare strings which in particular sentences can serve both as overlaps in an ambiguous sequence of phrases. Thus, when they achieve good scores as overlaps the number of phrases they participate increases and this makes their scores drop in the next phase. However, when they are not active the  part of the phrases where they are suffixes and prefixes become optimal and they boost the probabilities of these rare strings implying a vicious circle.  

\begin{figure}
\includegraphics[width=1\textwidth]{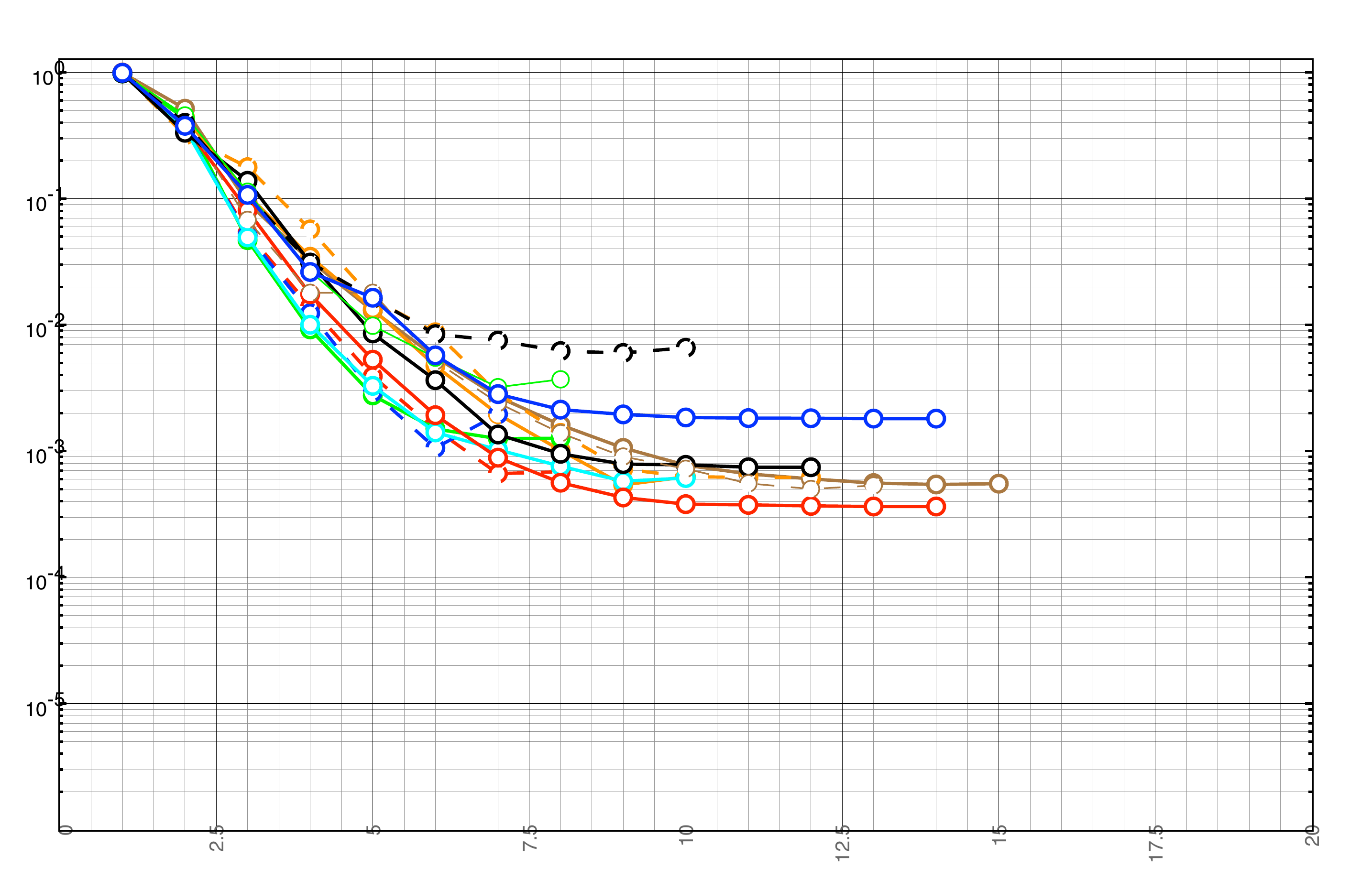}
\caption{Near-convergence of the algorithm for computing phrases. Behaviour of the similarity measure $\rho_i$ for the symmetric difference of phrase multisets for distinct rounds $i$ for corpora of different size, language and domain in logarithmic scale: 
{\bf\color{blue}---} Accountant ($\sim 2M$ Dutch sentences, newspaper articles); {\bf\color{cyan}---} excerpt of EU-parliament ($\sim 20 K$ German sentences, politics); {\bf\color{orange}---} excerpt of EU-parliament ($\sim 20 K$ French sentences, politics);
 {\bf\color{green}---} excerpt of EU-parliament ($\sim 20 K$ English sentences, politics);{\bf\color{red}---}excerpt of EU-parliament ($\sim 20 K$ Spanish sentences, politics);{\bf\color{purple}---} excerpt of EU-parliament ($\sim 20 K$ Dutch sentences, politics);{\bf\color{brown}---} excerpt of Medline ($\sim 15M$ English sentences, medicine); {\bf\color{yellow}---} excerpt of Medline ($\sim 800 K$ English sentences, medicine);{\bf\color{black}---} Wittgenstein ($\sim 4K$ German and English remarks, philosophy); ({\bf\color{black} - - -}) excerpt of the German Wikipedia Dump May 2014 ($\sim 2M$ sentences); ({\bf\color{green} - - -}) excerpt of the English Wikipedia Dump May 2014 ($\sim 2M$ sentences).}\label{fig:rho}
\end{figure}

This is confirmed in our experiments,  see Figure~\ref{fig:rho}, where we depict in logarithmic scale the behaviour of $\rho_i$ for distinct datasets. Multiset similarities $\rho_i$ start with values around $1$ and drop very quickly to values below $0.01$. Afterwards values $\rho_i$ slowly decrease to stabilise at rates $0.2\times 10^{-3}$ and $0.5\times 10^{-2}$, but they do not continue to converge to $0$.
Nevertheless, the multisets $\mathcal{P}_i$ exhibit a rather stable behaviour. In fact, more than $90\%$ of the value $\rho_i$ is due to cases where the multiplicity of a phrase $P$ is increased or decreased by just 1,  
$|m_i(P)-m_{i+1}(P)|=1$. Further the weighted {\em signed} symmetric difference between $\mathcal{P}_i$ and $\mathcal{P}_{i+1}$: 
\begin{equation*}
\delta_i=\frac{ \sum_{P} m_i(P)-m_{i+1}(P)}{\|\mathcal{P}_i\cup \mathcal{P}_{i+1}\|}.
\end{equation*} 
ranges at rates $|\delta_i|\in [10^{-6};10^{-4}]$ and below, which shows that typically a ``small gain'' for one phrase is compensated by a ``small loss'' of another phrase. This means, that on average, the changes between $p^{(i+1)}_{pref}$ and 
$p^{(i)}_{pref}$  (resp. $p^{(i+1)}_{suf}$ and $p^{(i)}_{suf}$) are small for function words with high probabilities. This shows that function 
words with high probabilities $p_{fw}^{(i)}$ do not change significantly their probabilities to $p_{fw}^{(i+1)}$. Whenever such a function word is admissible in a particular decomposition, it will be selected for a split. ``Persistent turbulences'' are caused by words occurring only once, or words occurring in a very specific syntactical contexts.

{\bf Technical realization.\ }
Technically, the set of all general phrase candidates $\mathcal{G}(\mathcal{C})$ can be represented after a linear preprocessing in linear space as a compressed acyclic word graph,~\cite{BlumerBlumer87,Inenaga01}, which generalises suffix trees,~\cite{Ukkonen95}. In this structure the required statistics, $\pref$, $\suff$, $\INF$ and $ov$ can be easily collected by a straightforward traversal. Meanwhile, the optimisation required in Equation~\ref{eq:likelihood} reduces
to a standard dynamic programming scheme. The graph necessary for this computation can be constructed by the means of an Aho-Corasick-like algorithm,~\cite{AhoCorasick75}. Further speed-up can be achieved by an A-star algorithm,~\cite{HNR68,HNR72}, built on top of the shortest path problem algorithm in proper interval graphs,~\cite{Atallah93}.

\subsection{Experimental basis - languages and corpora}
We considered corpora of different language, domain and size, see Table~\ref{table:corpora_data}. The corpora cover six languages: English, German, French, Spanish, Dutch, and Chinese.
Domains covered vary from philosophy through scientific medical abstracts, politics and journalistic articles, to general mixed topics. The size of the corpora scaled from several thousand to several Million sentences.

{\bf Wittgenstein Corpus.} The Wittgenstein Corpus consists of 3871 original remarks of the German philosopher Ludwig Wittgenstein. Although the general language is German, the corpus also contains a few remarks in English.

{\bf Medline Corpus.} The Medline Corpus is a collection of about 15 Million medical abstracts in English. Each abstract consists of about 10 sentences.
We considered two subsets of this corpus. They were obtained by randomly selecting 0.5\% and 10\%, respectively, of the abstracts and gathering all the sentences in the resulting abstracts. The 0.5\%-Medline Corpus consists of 800 K sentences, whereas the 10\%-Medline Corpus consists of about 15 M sentences.

{\bf EU-Parliament Corpus.} The EU-Parliament Corpus is a collection of the political statements in EU-Parliament from 1997 through 2012. The statements are maintained in all the languages of the EU-members at the date of the statement. Thus the entire corpus amounts to ca. 2 Million sentences  for the languages of the long-lasting EU-members, English, German, etc. and to several hundreds of thousands for the more recent EU-members.

From this corpus we selected excerpts from the 2000-2001 sessions of the EU-Parliament in English, German, French, Spanish and Dutch. The size of the selected corpora amounts to 20 K sentences. We also processed the entire French and Spanish corpora, amounting at about 2 Million sentences, each.

{\bf Accountant Corpus.} Accountant Corpus is a collection of the issues of the Dutch newspaper ``Accountant''. The texts are from  the period of World War I. The digital version of the corpus was obtained through OCR. 
We considered as \emph{sentences} sequences of characters separated by at least two new lines. Single new lines were replaced by white spaces.
In this way we tried to roughly reflect the natural reading order of paper articles. The obtained corpus, Accountant, contains  about 500 K different sentences (of total 780 K) most of which have the structure of a paragraph or article.

{\bf Wikipedia.} We considered also dumps of the English and German Wikipedias from May 14, 2014. 
In the English version we selected the pages AA -- AG in their raw form. In the German Corpus we selected the pages AA -- AH. We used WIKIPEDIAEXTRACTOR ~\cite{attardi2013} to remove the hyperlinks. In both cases the sentences were defined as sequences of characters terminated by a new line. In this way we roughly reflect the paragraph structure of these electronic resources. The size of the English corpus amounted to 1,1 Million paragraphs of which more than 890 K different and the German Corpus resulted in about 650 K paragraphs, more than 510 K of which were distinct.

{\bf Chinese Corpus.}  ``\begin{CJK*}{UTF8}{gbsn}保镖天下\end{CJK*}'' ("Bodyguard Legend")   is a Chinese electronic fiction which is freely available online  for research,~\cite{CHCorpus}. It contains about 24K of small paragraphs or single sentences. The total number of characters is about 2M.


\begin{table}
{\small
\begin{tabular}{|c|c|c|c|c|}\hline
Corpus & Lang. & Domain & Size & Size \\
&&&(sent.) & (words) \\ \hline
Wiki EN (May 2014) &  EN& General & $\sim 1,1 M$ & $\sim60 M$\\
&&&(paragraphs)& \\\hline
Wiki DE (May 2014) & DE & General & $\sim 650 K$ & $\sim24 M$\\\hline
&&&(paragraphs)& \\\hline
Accountant & NL & Newspaper articles & $\sim 780K$ & $\sim30 M$ \\
&&&(paragraphs)& \\\hline
Medline 10\% & EN & Medicine/Science & $\sim 15 M$ & $\sim292 M$ \\ \hline
Medline 0.5\% &EN & Medicine/Science & $\sim 800K$ & $\sim17M$ \\ \hline
EU-Parl FR 100\%& FR & Politics & $\sim 2M$ & $\sim52 M$ \\ \hline 
EU-Parl ES 100\%& FR & Politics & $\sim 2M$ & $\sim51 M$ \\ \hline 
EU-Parl EN 1\% & EN & Politics & $\sim 20 K$ & $\sim535 K$ \\ \hline
EU-Parl DE 1\%& DE & Politics & $\sim 20 K$ & $\sim531 K$ \\ \hline
EU-Parl FR 1\%& FR & Politics & $\sim 20 K$ & $\sim630 K$ \\ \hline
EU-Parl ES 1\%& ES & Politics & $\sim 20 K$ & $\sim593 K$ \\ \hline
EU-Parl NL 1\%& NL & Politics & $\sim 20 K$ & $\sim577 K$ \\ \hline
Wittgenstein & DE+EN & Philosophy & $\sim 4K$ remarks & $\sim250 K$ \\ \hline
Chinese & ZH & Fiction & $\sim 24K$ small   & $\sim 2M$ \\ 
 &  &  & paragraphs  &  characters\\ \hline
\end{tabular}}\caption{Metrics of the corpora considered in our experiments.}\label{table:corpora_data}
\end{table}

\subsection{Results on function words}
In this section we comment on the function words, $\mathcal{F}(\mathcal{C})$, that the algorithm described in
the previous sections detected in the various corpora. Table~\ref{table:all_delimiters} 
gives an overview for the European languages and distinct corpora. For each corpus we present the number of different
strings in $|\mathcal{F}(\mathcal{C})|$,  and we present a top segment of the function words. At this point, the order of
the function words is determined by the number of times a particular string has occurred as an overlap in an optimal decomposition. 
The corresponding results for the Chinese corpus are illustrated at the end of this subsection. 

Speaking in general terms, the number of the detected function words varies between 100--300. The results in Table~\ref{table:all_delimiters}
suggest that the actual number of function words does not depend on the size of the corpus, rather on its domain. For instance, the two Medline excerpts have a similar number of function words, although the size of one corpus is 20 times the size of the 
other. It is expected that in corpora with paragraph-like structure the number of function words increases. This is due to binding words occurring in the beginning of sentences which in a sentence-based corpus could not be used as overlaps. The ratio of the corpus size to the number of corpus units also exerts certain role.

Further, in a domain-specific corpus, such as EU-Parliament or Medline, we may expect that certain words usually considered as content words will often occur at the boundaries of phrases. For example\footnote{Here and in what follows we use the symbol \_ to highlight an occurrence of the white-space symbol.}, this is the case with \texttt{\_patients\_with\_} in the larger Medline corpus and with \texttt{\_patients\_} and \texttt{\_cell\_} in the smaller one. A manual inspection of the results reveals that there are 10-20 such strings in the domain-specific corpora which have the above property, yet only one or two in the Wikipedia corpora, i.e. \texttt{\_million\_} and \texttt{\_John\_}.

Besides the particular cases mentioned above, the obtained lists comprise of white space, punctuation symbols, articles, prepositions, pronouns, 
auxiliary verbs. A significant number of function ``words'' are multi-tokens such as ``to the''. 
All the function words in the Medline corpora and in most EU-Parliament corpora (EN, DE, NL and the smaller ES and FR) are delimited by white-spaces or punctuation symbols. There is a minimal number of exceptions in the other corpora (two in the English Wikipedia, one in the larger Spanish, and three exceptions in the larger French corpus, five in the Dutch Accountant Corpus and German Wikipedia and two in the Wittgenstein Corpus). The reason for these exceptions, e.g. ``\texttt{e.\_}'', ``\texttt{es.\_}'', ``\texttt{s.\_}'' in French or ``\texttt{en.\_}'' in German (Wittgenstein Corpus) and Dutch (Accountant), ``\texttt{s"\_}'' in the English Wikipedia, can be explained in the following way. Some classes of words are underrepresented in the context of specific punctuation symbols. For instance, in German relative clauses often end with a verb with suffix \texttt{en}. Thus, we often observe ``\texttt{en.\_}'' at the end of the sentence. Yet, considering a particular verb form it is by no means evident that it will occur twice at the end of a sentence.

\begin{table}
\begin{center}
\caption{Function words extracted from English, German, Dutch, Frence, and Spanish corpora: number of detected function words, $\vert\mathcal{F}(\mathcal{C})\vert$, and function words most often used 
in decompositions.}\label{table:all_delimiters}
\end{center}
{\small 
\begin{tabular}{|c|c|l|}\hline
Corpus & $|\mathcal{F}(\mathcal{C})|$ & function words used most often \\\hline
WikiEN & 300 & \_; the ;\texttt{.} ; and ; of ; in ; to ; a ; of the ; is ;\texttt{"} ; was ;\texttt{.} The ; in the ;  \\ && 
for ; that ; by ; as ; with ; are ;\texttt{-};\texttt{(} ; on ; were ; to the ; from ; at ; \\ && his ;\texttt{.} In ;\texttt{",} ;\texttt{;} ;  an ; or ; which ;\\
\hline
EU-Parl & 198 & \_; the ; \texttt{,} ; to ; of ; a ; and ; in ; that ; be ; for ; this ; which ; in the ; \\
(EN) 1\%& &  are ; we ; will ; have ;  on ; has ; an ; not ; to the ; with ; \texttt{,} the ;  \\ && that the ; it ; as ; by ; to be ; \texttt{,} and ; more ; \\
\hline
Medline & 145 & \_; of ; and ; the ; in ; of the ;  to ; was ;  with ; for ;  a ;  were ; is ;  in the ; \\
(10\%) & &  by ; that ; are ;)\texttt{,} ; from ; on ; at ; or ; after ; as ;  +/- ; and the ;  \\ && between ; during ; to the ;\texttt{;} ; patients with ; \\ 
\hline
Medline & 143 &  \_; of ; the ; and ; in ; to ; of the ; was ; a ; were ; with ; for ; is ; in the ;   \\ 
(0.5\%) & & by ; that ; are ; from ;)\texttt{,} ; at ; on ;  as ; after ; to the ; or ; an ;  \\ &&  patients ;  +/- ; cells ;\texttt{;} ; and the ;
\\ \hline
%
\hline \ 
WikiDE & 302 & \_;\texttt{.} ;  der ; die ;  und ; des ; in ; von ; \texttt{"}; -; im ; das ;\texttt{,} ;\texttt{)} ; eine ; den ; \\
&& als ;\texttt{.} Die ; zu ;  mit ; ein ; ist ;\texttt{)} in der ;  wurde ; f\"ur ; auf ; zur ; einer ; \\ &&  nach ;  sind ;\texttt{.} Der ; nicht ;  \\
\hline
EU-Parl 1\%& 158 & \_; \texttt{,} ; die ; der ; und ; zu ; in ; den ; eine ;\texttt{,} da\ss ; des ; \texttt{,} die ; das ; \\
(DE)  &  &  nicht ; von ;  werden ; im ; ist ; ein ; f\"ur ; wir ; auf ; es ; einer ;  auch ;\\ &&  zur ; mit ; dem ; sich ;  
\\
\hline
Wittgenstein\footnote{In this experiment we used as threshold $|\mathcal{C}|/300$ instead of $|\mathcal{C}|/1000$. This is motivated by the fact that the corpus is much smaller in terms of $|\mathcal{C}|<4000$.} & 173 &  \_;\texttt{.} ;[; die ; der ; das ;\texttt{[1]};\texttt{?} ; the ; \& ; da\ss ; nicht ; eine ; ein ; ist ; \\
&& den ; to ; es ;\texttt{,} ;  zu ; \texttt{;} ; a ; so ; von ; of ; that ; wenn ; was ;  I ; 
\\ &&  er ; als ; einen ; wie ; aber ; sie ;
\\ \hline
%
\hline  \ 
Accountant & 373 & \_; de ; een ;en. ;\texttt{.} ; en ; in ; dat ; is ;\texttt{,} ; die ; \texttt{.} De ; \texttt{(}; niet ; op ;\texttt{:} ;\\
&& \texttt{;} ;in de ; van het ; zijn ; voor ;\texttt{\^}; met ;  als ; dat de ;\texttt{.} Het ; wordt ;  
 \\
\hline
EU-Parl& 151 & \_; \texttt{, }; van ; een ; het ; in ; en ; van de ; dat ; te ; op ; die ; is ; niet ; \\
(NL)  1\%&&  voor ;  zijn ; de ; worden ;  met ; om ; dat de ; deze ; heeft ; van het ;  \\ && aan ; voor de ; ook ; er ; wij ; 
\\
\hline
%
\hline
EU-Parl & 145 & \_;\texttt{,} ; de ; des ; les ; \`a ; que ; et ; une ; un ; en ; du ; qui ; pour ; pas ; \\   
(FR) 1\%&& dans ; nous ; au ; a ; il ; aux ; plus ; que la ; sur ; se ; par ; \^etre ; d ' ;  
\\ \hline
EU-Parl& 198 & \_;\texttt{,} ; de ; des ; et ; les ; la ; \`a ; que ; en ; pour ; une ; du ; qui ; dans ; \\
(FR)  100\%&& un ; au ; est ;  aux ; a ; \'a la ; sont ; que les ;\texttt{,} et ;\texttt{,} mais ; par ; ont ; 
\\ && sur ; d'une ;\texttt{,} les ; plus ; d'un ; 
\\ 
\hline
%
\hline
EU-Parl & 140 & \_ ;\texttt{,} ; de ; que ; en ; el ; y ; a ;  los ;  del ; una ; las ; un ; se ; para ;  \\
(ES) 1\% && de los ; no ; con ; es ;  al ; a la ; ha ;de las ; por ; la ; su ; que la ; m\'as ; 
\\
\hline
EU-Parl & 172 & \_ ;\texttt{,} ; que ; la ; en ; y ; los ; el ; a ; para ; del ; una ; se ; un ; las ; con ; \\
(ES) 100\% &&  no ; al ; a la ; es ; por ; en la ; de las ; a los ;\texttt{,} que ;\texttt{,} y ; ha ; m\'as ; su ;  \\ && sobre ; como ;\texttt{:} ;\texttt{,} pero ; y la ; a las ;
\\
\hline
\end{tabular}}
\end{table}

{\bf Function words obtained for the ``\begin{CJK*}{UTF8}{gbsn}保镖天下\end{CJK*}'' corpus.\ }  
Since in Chinese the words of a sentence are not delimited it was interesting to see how the approach would work in this situation. Remarkably, results do not strongly differ from the other cases.  For our Chinese corpus, ``\begin{CJK*}{UTF8}{gbsn}保镖天下\end{CJK*}'', which represents fiction, we obtained a total of 291 function words. 
We present the top 50 function words most often used in the decompositions with accompanying English equivalent. 
``'' (empty string);\quad ``\begin{CJK*}{UTF8}{gbsn}，\end{CJK*}'' (comma white space);\quad 
``\begin{CJK*}{UTF8}{gbsn}的\end{CJK*}'' (of);\quad
``\begin{CJK*}{UTF8}{gbsn}了\end{CJK*}'' (over);\quad
``\begin{CJK*}{UTF8}{gbsn}是\end{CJK*}'' (is/are);\quad
``\begin{CJK*}{UTF8}{gbsn}你\end{CJK*}'' (you);\quad
``\begin{CJK*}{UTF8}{gbsn}我\end{CJK*}'' (I/me);\quad
``\begin{CJK*}{UTF8}{gbsn}这\end{CJK*}'' (this);\quad
``\begin{CJK*}{UTF8}{gbsn}人\end{CJK*}'' (people/human);\quad
``\begin{CJK*}{UTF8}{gbsn}着\end{CJK*}'' (no specific meaning in Chinese);\quad
``\begin{CJK*}{UTF8}{gbsn}在\end{CJK*}'' (at/in);\quad
``\begin{CJK*}{UTF8}{gbsn}就\end{CJK*}'' (no specific meaning in Chinese);\quad
``”'' (apostrophs);\quad
``\begin{CJK*}{UTF8}{gbsn}黎箫\end{CJK*}'' (Li Xiao -- a person's name in this fiction);\quad
``\begin{CJK*}{UTF8}{gbsn}他\end{CJK*}'' (he);\quad
``\begin{CJK*}{UTF8}{gbsn}也\end{CJK*}'' (also/too);\quad
``\begin{CJK*}{UTF8}{gbsn}到\end{CJK*}'' (arrive);\quad
``\begin{CJK*}{UTF8}{gbsn}？\end{CJK*}'' (?);\quad
``\begin{CJK*}{UTF8}{gbsn}都\end{CJK*}'' (both/all);\quad
``\begin{CJK*}{UTF8}{gbsn}去\end{CJK*}'' (go);\quad
``\begin{CJK*}{UTF8}{gbsn}！\end{CJK*}'' (!);\quad
``\begin{CJK*}{UTF8}{gbsn}来\end{CJK*}'' (come);\quad
``\begin{CJK*}{UTF8}{gbsn}还\end{CJK*}'' (still/again);\quad
``\begin{CJK*}{UTF8}{gbsn}看着\end{CJK*}'' (look/stare);\quad
``\begin{CJK*}{UTF8}{gbsn}~！\end{CJK*}'' (~!);\quad
``~\begin{CJK*}{UTF8}{gbsn}！”\end{CJK*}'' (~!);\quad
``\begin{CJK*}{UTF8}{gbsn}那\end{CJK*}'' (that);\quad
``\begin{CJK*}{UTF8}{gbsn}一个\end{CJK*}'' (one);\quad
``\begin{CJK*}{UTF8}{gbsn}出\end{CJK*}'' (go/get out);\quad
``\begin{CJK*}{UTF8}{gbsn}上\end{CJK*}'' (up/above);\quad
``\begin{CJK*}{UTF8}{gbsn}要\end{CJK*}'' (need);\quad
``\begin{CJK*}{UTF8}{gbsn}和\end{CJK*}'' (and);\quad
``\begin{CJK*}{UTF8}{gbsn}：“\end{CJK*}'' (:apostrophs)\quad
``\begin{CJK*}{UTF8}{gbsn}自己\end{CJK*}'' (myself);\quad
``\begin{CJK*}{UTF8}{gbsn}将\end{CJK*}'' (will/shall);\quad
``\begin{CJK*}{UTF8}{gbsn}没有\end{CJK*}'' (no/didn't);\quad

`` \begin{CJK*}{UTF8}{gbsn}能\end{CJK*}'' (can/be able to);\quad
``\begin{CJK*}{UTF8}{gbsn}后\end{CJK*}'' (behind/later);\quad
``\begin{CJK*}{UTF8}{gbsn}被\end{CJK*}'' (be--used in passive tense, before a verb);\quad
``\begin{CJK*}{UTF8}{gbsn}会\end{CJK*}'' (will/can);\quad
``\begin{CJK*}{UTF8}{gbsn}向\end{CJK*}'' (towards);\quad
``\begin{CJK*}{UTF8}{gbsn}已经\end{CJK*}'' (already);\quad
``\begin{CJK*}{UTF8}{gbsn}让\end{CJK*}'' (let);\quad
``\begin{CJK*}{UTF8}{gbsn}对\end{CJK*}'' (write/correct);\quad
``\begin{CJK*}{UTF8}{gbsn}却\end{CJK*}'' (but/and yet/however);\quad
``\begin{CJK*}{UTF8}{gbsn}给\end{CJK*}'' (give/provide);\quad
``\begin{CJK*}{UTF8}{gbsn}竟然\end{CJK*}'' (unexpectedly/to one's surprise);\quad
``\begin{CJK*}{UTF8}{gbsn}我们\end{CJK*}'' (we/us);\quad
``\begin{CJK*}{UTF8}{gbsn}一脸\end{CJK*}'' (the whole face);\quad
``\begin{CJK*}{UTF8}{gbsn}现在\end{CJK*}'' (now/currently);\quad

 \subsection{Quantitative results on phrases}
In this section we provide some quantitative statistics on the multiset of phrases $\mathcal{P}$ extracted in the final step of our algorithm. 
From our point of view it is interesting to assess how complex the resulting phrases are and how often they have been selected by the algorithm
to obtain an optimal decomposition of a sentence.

To assess the first characteristic of $\mathcal{P}(\mathcal{C})$ we counted in a {\em postprocessing} step the number of words, $word(P)$, contained in each of the 
strings $P\in \mathcal{P}(\mathcal{C})$. This was done by counting the number of white spaces, $ws(P)$, in each of the strings $P\in\mathcal{P}(\mathcal{C})$ and subtracting 
one, unless the string itself turned out to be a beginning or an end of a sentences in the corpus. Thus, $ws(P)$ takes on values $0,1,\dots$ and 
$word(P)$ takes on values $-1,0,\dots$. The values $word(P)=-1$ and $word(P)=0$ mean that the string contains no word. We found out that strings 
of this form are unavoidable, details are left out.

\begin{figure}
\includegraphics[width=0.8\textwidth]{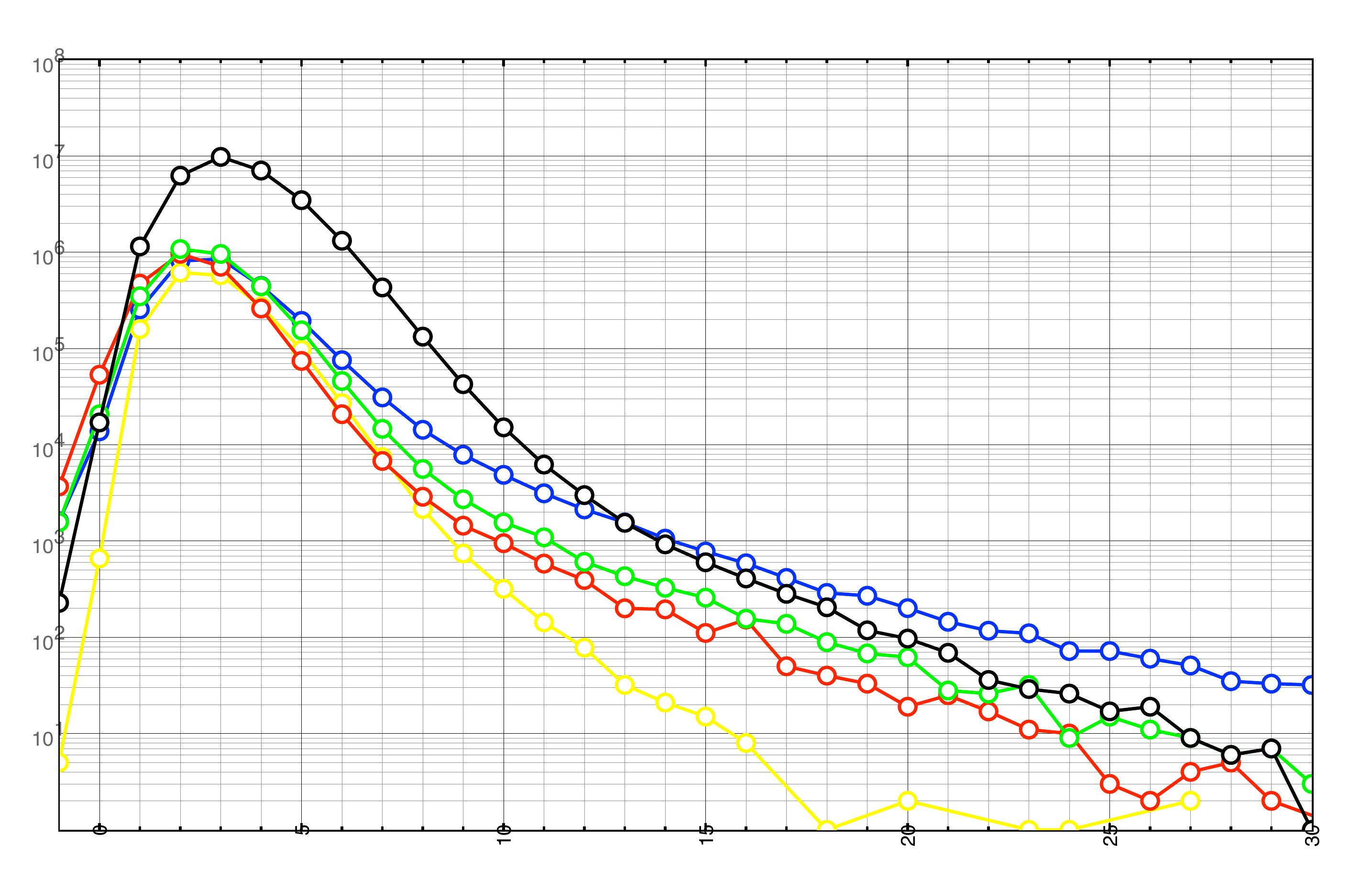}
\includegraphics[width=0.8\textwidth]{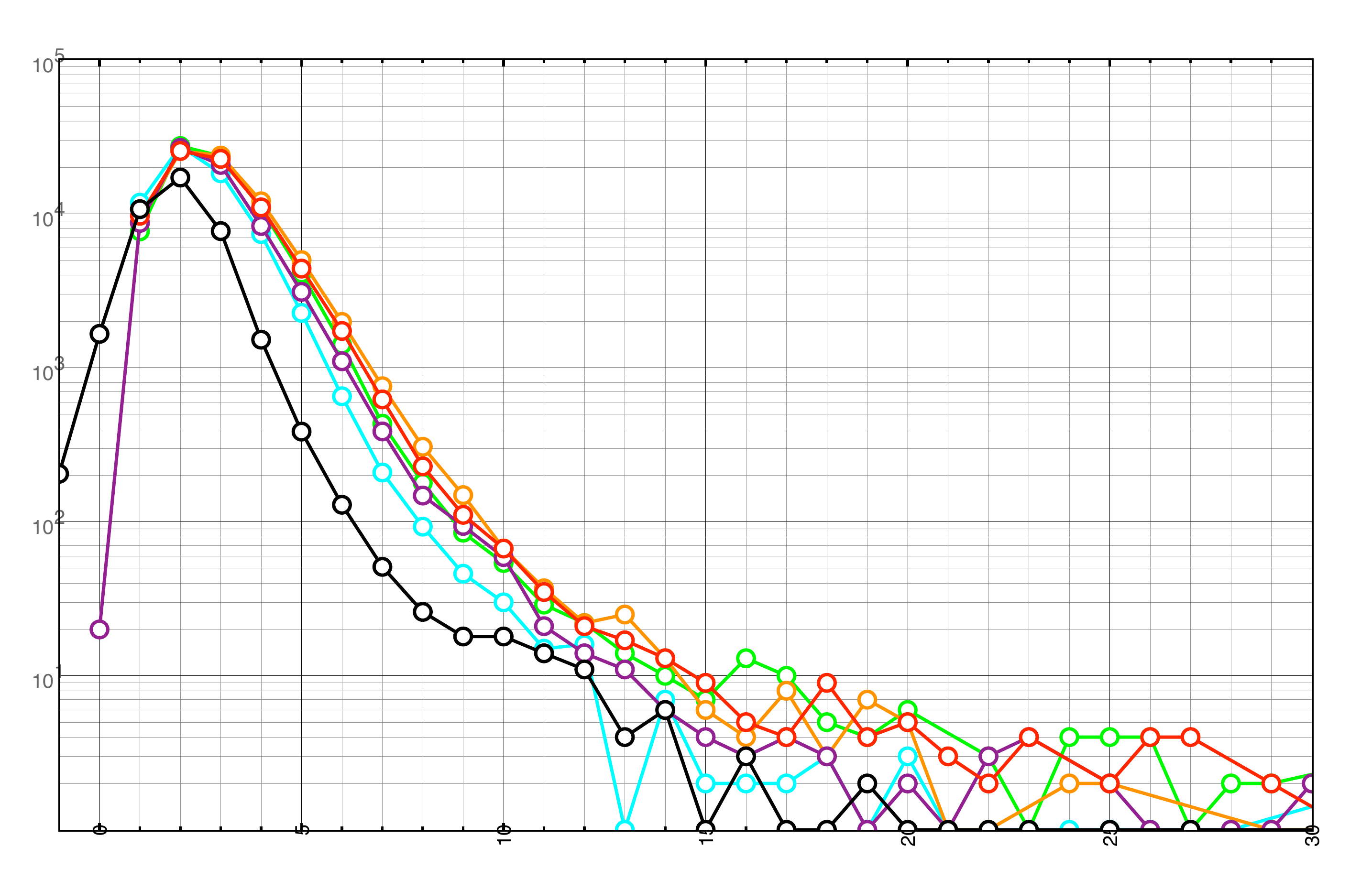}
\caption{Distribution of the number of phrases containing $n$ words: $word(P)\le 30$ against the logarithmic scale of the number of phrases $P\in\mathcal{C}(\mathcal{P})$: Upper diagram:({\bf\color{black}---}) excerpt of Medline Corpus ($\sim 15M$ sentences); ({\bf\color{yellow}---}) excerpt of Medline Corpus ($\sim 800K$ sentences); ({\bf\color{blue}---}) Accountant Corpus ($\sim 2M$ sentences); ({\bf\color{red}---}) excerpt of German Wikipedia Dump May 2014 ($\sim 2M$ sentences, $5\%$ of the entire Dump); ({\bf\color{green}---}) excerpt of English Wikipedia Dump May 2014 ($\sim 2M$ sentences, $2\%$ of the entire Dump).
Lower diagram: EU-Parliament Corpora excerpt ($\sim 20K$ sentences) in German ({\bf\color{cyan}---}); English ({\bf\color{green}---}); French ({\bf\color{orange}---}); Dutch ({\bf\color{pink}---}); Spanish ({\bf\color{red}---}); and the Wittgenstein Corpus ($\sim 4K$) remarks ({\bf\color{black}---}).
}\label{fig:big_corpora30}\label{fig:big_word_clean30}
\end{figure}

Figure~\ref{fig:big_word_clean30} 
shows the distribution of the number of phrases in $\mathcal{P}(\mathcal{C})$ w.r.t. their $word(P)$-lengths for $word(P)\le 30$. The variety of longer phrases $word(P)=2,3,4$ is larger than the variety of single-word phrases. It is worth to note that number of cases $word(P)=-1,0$ is very small. Manual inspection of such cases shows that these strings result from some particular features of the corpus, e.g. special symbols such as ``{\verb+^+}'' in the OCR-ed corpus ``Accountant'', special nomenclature and indices of the remarks in the Wittgenstein Corpus, chemical formulas in the Medline Corpus etc.

\begin{figure}
\includegraphics[width=0.8\textwidth]{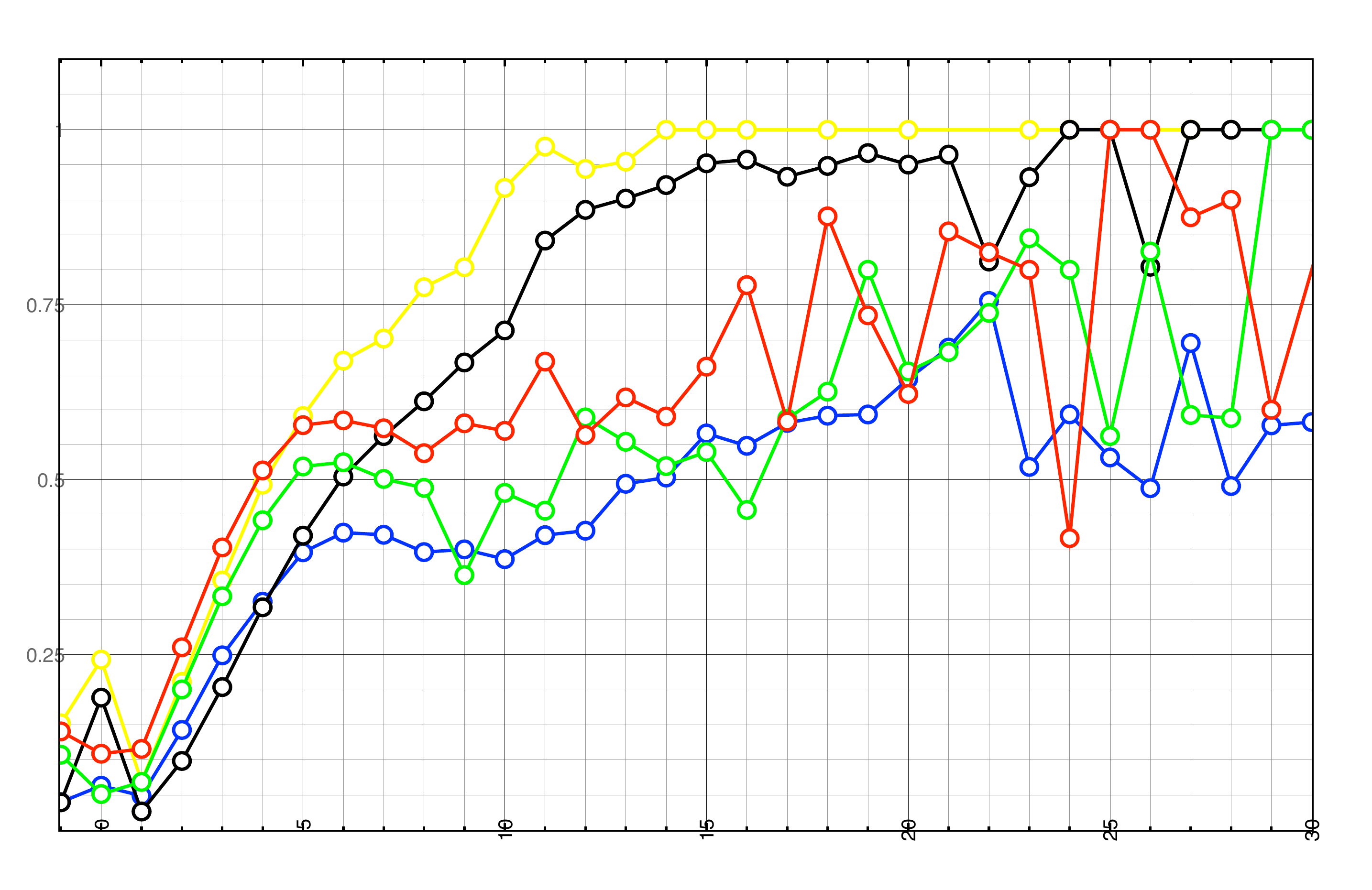}
\includegraphics[width=0.8\textwidth]{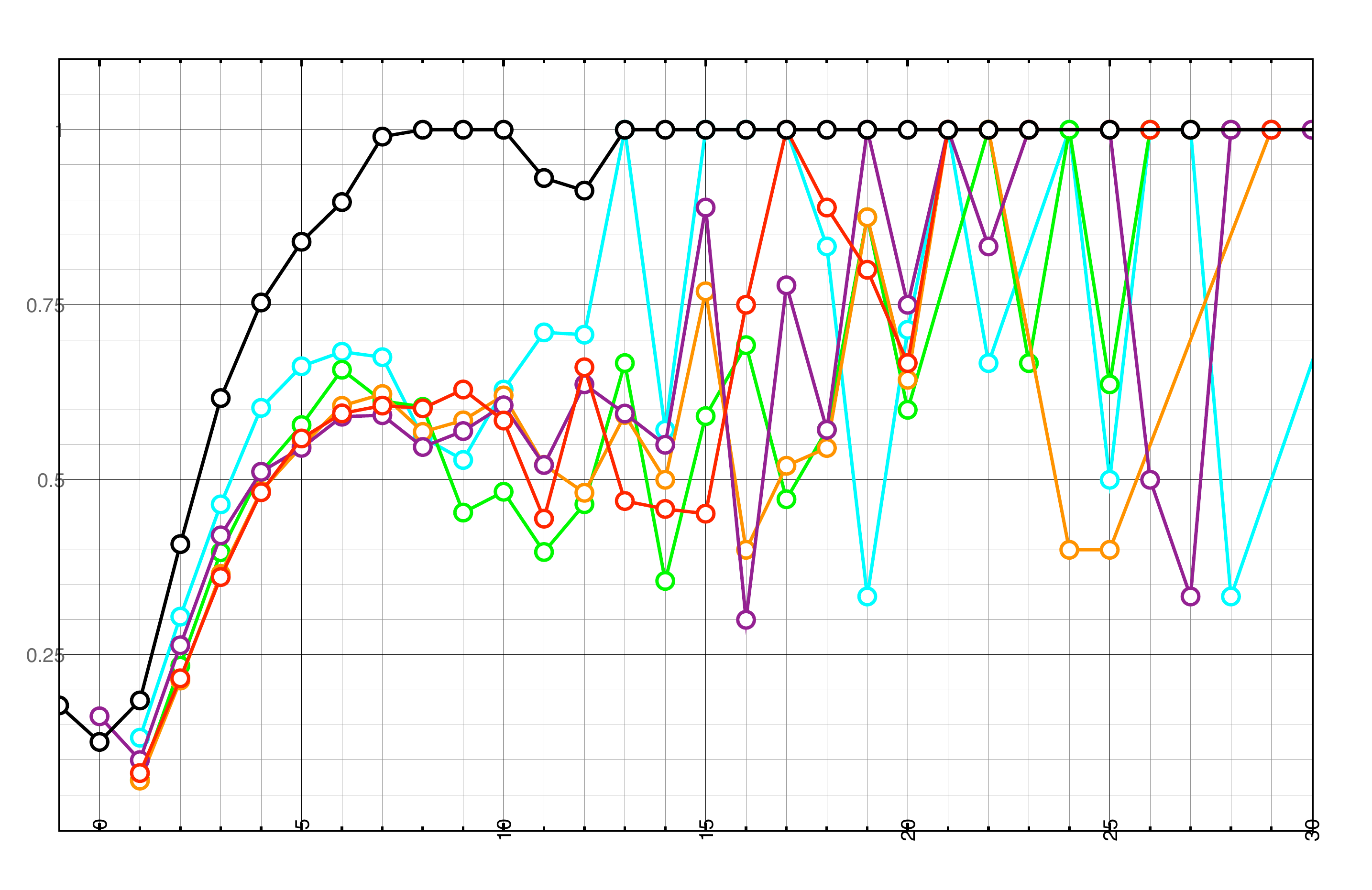}
\caption{How often phrases are used for sentence decomposition - dependency on the number of words in the phrase:  $w\le 30$ (x-axis) and $M_{\mathcal{C}}(w)/O_{\mathcal{C}}(w)$ for the excerpt of Medline Corpus ($\sim 800K$ sentences) and the Accountant Corpus ($\sim 2M$ sentences) (upper diagram), and EU-Parliament Corpora excerpt ($\sim 20K$ sentences) in German ({\bf\color{cyan}---}); English ({\bf\color{green}---}); French ({\bf\color{orange}---}); Dutch ({\bf\color{pink}---}); Spanish ({\bf\color{red}---}); and the Wittgenstein Corpus ($\sim 4K$) remarks ({\bf\color{black}---}) (lower diagram).}\label{fig:big_corpora30}\label{fig:big_prefer_clean30}
\end{figure}

In order to estimate phrases of what length were actually preferred by the algorithm we proceeded in the following way. For $w=-1,0,\dots$
we computed the total number of multiplicities, $M(w)$, of a string $P\in\mathcal{P}(\mathcal{C})$ with $word(P)=w$, i.e.:
\begin{equation*}
M(w)=\sum_{P\in\mathcal{C}:word(P)=w} m_n(P).
\end{equation*}
We also computed the total number of occurrences, $O(w)$, in the corpus of the strings $P\in\mathcal{P}$ with $word(P)=w$, i.e.:
\begin{equation*}
O(w)=\sum_{P\in\mathcal{C}:word(P)=w, m_n(P)\ge 1} occ(P).
\end{equation*}
Thus, the ratio $M(w)/O(w)$ provides an average on which percentage of the occurrences of a phrase $P\in \mathcal{P}$ with $word(P)=w$ has been  used in the final decompositions determined by $\mathcal{P}$.
In Figure~\ref{fig:big_prefer_clean30} 
we depict the relationship between $w$ and $M(w)/O(w)$ for $w\le 30$ for the corpora w.r.t. their size. Results confirm that the algorithm selects longer phrases with higher preference than shorter ones.

\subsection{Qualitative results on phrases}\label{SubsectionQualitative}

In our approach, the set of phrases represents a subset of the set of general phrase candidates, $\mathcal{G}(\mathcal{C})$. The above conditions imposed on general phrase candidates imply that a
string with only one occurrence in $\mathcal{C}$ that does not represent a full sentence cannot belong to $\mathcal{G}(\mathcal{C})$. 
This means that only those proper phrases can be found that have at least two occurrences (with distinct immediate contexts) in the given corpus $\mathcal{C}$. Since long substrings often have just one occurrence, many long substrings that we would treat - from a linguistic point of view - as phrases are not found in ${\cal P}({\cal C})$. As a matter of fact, we cannot obtain another result when we demand that phrases must occur in distinct contexts in the given corpus. 
From a qualitative point of view, the essential characteristics of the set of extracted phrases 
$\mathcal{P}(\mathcal{C})$ can be summarized in the following way. Examples for all cases are presented below. 
\begin{enumerate}
\item 
Ignoring a small percentage of exceptions, elements $P \in \mathcal{P}(\mathcal{C})$ 
are sequences of ``words'' delimited by blanks or special sentence symbols (,.;:). At this point it should be kept in mind that our algorithm does not know the notion of a word, nor the role of distinct letters in the alphabet. Exceptions where elements $P \in \mathcal{P}(\mathcal{C})$ are not delimited by blanks (or sentence symbols) are caused by strings with only one occurrence in the corpus. 
\item
In most cases, elements $P \in \mathcal{P}(\mathcal{C})$  extracted are meaningful units in the sense that they represent phrases, or 
phrases extended on the left/right border by characteristic overlapping units, function words. In addition we often find units that would be interesting for language learners, showing how kernel sequences of words are typically connected with neighboured phrases. For the sake of reference, such units will be called {\em phrases extended by typical connectors}. 
\item 
When ignoring delimiting functions words, linguistic phrases in the corpus in most cases are 
representable as elements in $\mathcal{P}(\mathcal{C})$ or sequences of such elements. 
\end{enumerate}

\begin{example}\label{ExDec1}
We present the decomposition of the Europarl-sentence \texttt{You have requested a debate on this subject  in the  course of the next few days , during this part-session .} (en) and the parallel sentences for the languages French (fr), 
Spanish (es), Dutch (du), and German (de). In the following sentence decompositions, phrases $P \in \mathcal{P}(\mathcal{C})$ extracted are the substrings marked by upper brackets $^{(}\ldots ^{)}$ or 
lower brackets $_{(}\ldots _{)}$, and overlaps of decompositions have the form $^{(}\ldots _{)}$ or $_{(}\ldots ^{)}$, function ``words'' in overlaps highlighted in bold. 
\begin{enumerate}
\item[(en)] \texttt{$^($You have$_($ $^)$requested a$^($ $_)$debate$_($ {\bf on this} $^)$subject$^($ {\bf in the} $_)$course of the next few days$_($ $^)$, during \\ this$^($ $_)$part-session .$^)$}
\item[(fr)] \texttt{$^($Vous avez$_($ $^)$souhait{\'e}$^($ $_)$un d{\'e}bat {\`a} ce sujet$_($ $^)$dans les prochains jours$^($ {\bf , }$_)$au cours$_($ {\bf de cette} $^)$p{\'e}riode de \\ session$^($ $_)$.$^)$}
\item[(es)] \texttt{$^($Sus Se{\~n}or{\'i}as han$_($ $^)$solicitado un$^($ $_)$debate sobre el\\ $_($ $^)$tema para$^($ $_)$los pr{\'o}ximos d{\'i}as ,$_($ $^)$en el curso$^($ {\bf de} $_)$\\ este per{\'i}odo$_($ $^)$de sesiones .$_)$}
\item[(du)] \texttt{$^($U heeft aangegeven dat u$_($ $^)$deze vergaderperiode$^($ $_)$\\ een debat$_($ $^)$wilt$^($ $_)$over deze$_($ $^)$rampen .$_)$}
\item[(de)] \texttt{$^($Im Parlament$_($ $^)$besteht der$^($ $_)$Wunsch nach$_($ {\bf einer} $^)$\\ Aussprache $^($ $_)$im Verlauf$_($ $^)$dieser Sitzungsperiode$^($ $_)$\\ in den n\"achsten Tagen$_($ $^)$.$_)$}
\end{enumerate}
\end{example}
Several long phrases  with not just one occurrences in the corpus are recognized. 
Examples are 
\begin{eqnarray*}
&&\texttt{\_in\_the\_course\_of\_the\_next\_few\_days\_}\\
&&\texttt{\_un\_d{\'e}bat\_{\`a}\_ce\_sujet\_} \\
&&\texttt{\_in\_den\_n\"achsten\_Tagen\_} 
\end{eqnarray*}
Looking at the sentence beginnings we find ``phrases extended by typical connectors'' in the above sense:
\begin{eqnarray*}
&&\texttt{You\_have\_}\\
&&\texttt{Vous\_avez\_} \\
&&\texttt{U\_heeft\_aangegeven\_dat\_u\_}\\
&&\texttt{Sus\_Se{\~n}or{\'i}as\_han\_} 
\end{eqnarray*}
Examples where long phrases can be represented as sequence of units in $\mathcal{P}(\mathcal{C})$ (possibly deleting overlaps on the two sides) are
\begin{eqnarray*}
&&\texttt{You\_have\_requested\_a\_debate\_on\_this\_subject}\\
&&\texttt{Vous\_avez\_souhait{\'e}\_un\_d{\'e}bat\_{\`a}\_ce\_sujet} \\
&&\texttt{au cours\_de\_cette\_p{\'e}riode\_de\_session\_} \\
&&\texttt{de\_este\_per{\'i}odo\_de\_sesiones\_.} \\
&&\texttt{\_im\_Verlauf\_dieser\_Sitzungsperiode\_}
\end{eqnarray*}

\begin{example}\label{ExDec2}
We present a typical decomposition of a sentence of the Medline Corpus.\\

\texttt{$^($A$_($ {\bf viral} $^)$etiology,$^($ $_)$eg, the$_($ $^)$congenital rubella syn-\\ drome, $^($ $_)$was considered most likely,$_($ $^)$but detailed$^($ $_)$\\ investigations$_($ {\bf proved to be} $^)$negative.$_)$}\\

\noindent In this sentence, ignoring extensions caused by the comma ``,'', terminological expressions such as 
\begin{eqnarray*}
&&\texttt{\_viral\_etiology}\\
&&\texttt{\_congenital\_rubella\_syndrome} 
\end{eqnarray*}
are found as phrases. In fact, since terminological expressions often appear in distinct contexts, one strength of the method is the ability to find terminological phrases (see Section~\ref{SubSec4.1}).  The noun phrase \texttt{detailed investigations} is split, at this point corpus characteristics cause the effect that  \texttt{\_but\_detailed\_} and \texttt{\_investigations\_proved\_to\_be\_} are prefered. Phrases found also include
\begin{eqnarray*}
&&\texttt{\_was\_considered\_most\_likely,\_}\\
&&\texttt{\_proved\_to\_be\_negative.} 
\end{eqnarray*}
Phrases that can be obtained by combining recognized units are, e.g.,
\begin{eqnarray*}
&&\texttt{\_but\_detailed\_investigations\_proved\_to\_be\_negative.} 
\end{eqnarray*}
Also note that \texttt{\_viral\_} and \texttt{\_proved\_to\_be\_} occur as an overlap, a characteristic feature of the Medline corpus. 
\end{example}

\begin{example}\label{ExDec3}
We present a typical decomposition of a sentence of the English Wikipedia.\\

\texttt{$^($She further$_($ $^)$demonstrated that$^($ $_)$much of the other$_($ $^)$\\ research in the book$^($ $_)$arguing for$_($ $^)$a link$^($ {\bf between} $_)$\\ Sir$_($ {\bf John} $^)$Williams and$^($ $_)$the Ripper crimes$_($ $^)$was flawed.$_)$}\\

\noindent Here \texttt{\_between\_} and \texttt{\_John\_} occur as an overlap. 
Phrases found are 
\begin{eqnarray*}
&&\texttt{\_research\_in\_the\_book\_}\\
&&\texttt{\_the\_Ripper\_crimes\_} 
\end{eqnarray*}
Interesting units that language learners would consider as important expressions, but 
not representing linguistic phrases (``phrases extended by typical connectors'' in the above sense), are 
\begin{eqnarray*}
&&\texttt{\_arguing\_for\_}\\
&&\texttt{\_a\_link\_between\_} 
\end{eqnarray*}
\end{example}

\begin{example}\label{ExDec4}
We present a typical decomposition of a sentence of the German Wikipedia.\\

\texttt{$^($vom$_($ $^)$Orchestre de la Suisse Romande$_)${$^($}, den$_($ $^)$Berliner \\ Philharmoniker,$^($ $_)$dem Tonhalle-Orchester Z\"urich$_($ $^)$mit\\ Werken $^($ $_)$klass$_(${\bf ischer Komponisten}$^)$, $^(${\bf wie} $_)$auch$_($ $^)$Komponisten\\ $^($ $_)$ Polens$_($ von $^)$Fr\'ed\'eric Chopin$^($ $_)$bis$_($ $^)$Witold Lutoslawski\\ $^($ $_)$\"uber$_($ $^)$Krzysztof Penderecki.$_)$}\\

\noindent Due to particularities of the corpus,  \texttt{\_ischer\_Komponisten\_} occurs as an overlap. 
Phrases directly found are 
\begin{eqnarray*}
&&\texttt{\_Orchestre\_de\_la\_Suisse\_Romande}\\
&&\texttt{\_Berliner\_Philharmoniker,\_}\\ 
&&\texttt{\_dem\_Tonhalle-Orchester\_Z\"urich\_} 
\end{eqnarray*}
Phrases corresponding to combinations of recognized units are 
\begin{eqnarray*}
&&\texttt{\_mit\_Werken\_klassischer\_Komponisten\_}\\
&&\texttt{\_von\_Fr\'ed\'eric\_Chopin\_bis\_Witold Lutoslawski\_} 
\end{eqnarray*}
Also note that person names are well recognized. This is again due to the effect that these person names appear in distinct contexts in the corpus. 
\end{example}

\begin{example}\label{ExDec5}
We present a typical decomposition of a sentence of the Chinese corpus. The sentence is  
\begin{quotation}
``\begin{CJK*}{UTF8}{gbsn}老者顺手递过来一份文件，黎箫打开一看，眼睛立刻就直了，没看别的，就看那张不大的照片，美女啊，绝对的美女啊，我们黎箫口水“飞流直下三千尺”\end{CJK*}''
\end{quotation}
the English translation is 
\begin{quotation}
``The old man handed over a document, LiXiao stared at it with motionless eyes as soon as he opened it, looking at nothing else, just the small picture, beautiful girl, a really beautiful girl, the saliva of our Lixiao has flowed three thousand feet.''
\end{quotation}
The sentence decomposition determined by our algorithm is the following.
\begin{quotation}
``\begin{CJK*}{UTF8}{gbsn}$^($老者$^)_($顺手$_)^($递过$_($来$^)$一份$_)^($文件$_($，$^)$黎箫打开$_)^($一看，眼睛立刻$^)_($就直了$^($，$_)$没看$^)_($别的$^($，$_)$就看$^)_($那张$_)^($不大$_($的$^)$照片$^($，$_)$美女$^)_($啊，绝对的$_)^($美女啊$_($，$^)$我们黎箫$_)^($口水$^)_($“飞$_)^($流$^)_($直下$_)^($三千$^)_($尺”$_)$\end{CJK*}''
\end{quotation}
It has the following (partially overlapping) phrases:
\begin{quotation}
``\begin{CJK*}{UTF8}{gbsn}$^($老者old man$^)_($顺手smoothly$_)^($递过来hand (something to somebody)$^)$ $_($来一份one piece$_)^($文件document, or paper，$^)_($, Lixiao opens黎箫打开$_)^($一看，眼睛立刻looks, eyes immediately$^)_($就直了motionless/steady，$_)$ $^($, 没看did not look$^)_($别的others，$_)^($, 就看only look$^)_($那张that piece of$_)^($不大的not big, small$^)_($的照片picture，$_)^($, 美女beautiful girl$^)_($啊a，绝对的absolute$_)^($美女啊beautiful girl，$^)_($, 我们黎箫Lixiao or our Lixiao$_)^($口水saliva$^)$ $_($“飞fly$_)^($流flow$^)_($直下straightly down$_)^($三千three thousand$^)_($尺”feet$_)$\end{CJK*}''
\end{quotation}

\end{example}

\subsection{Computing extended sets of phrases}\label{SubsectionExtensions}

An evident drawback of the approach proposed above is its incapability to detect phrases occurring only once: since these phrases do not appear in distinct contexts they are not contained in the 
initial set of general phrase candidates ${\cal G}({\cal C})$. Obviously, many substrings occurring only once represent interesting language units. The problem is challenging because we do not have immediate statistical evidence that helps to detect such substrings. We looked at two ways to address this problem, in both cases using a dynamic approach.


{\bf 1. Extended island phrases.\ } We consider the sentences, $S_n$,
in the natural order of their appearance in the corpus $\mathcal{C}$. At each time step, $t$, 
we let $\mathcal{C}_t$ be the set of sentences observed up to moment $t$.
Using a dynamic interpretation of assumptions \ref{lab_amb}-\ref{lab_str}, the set of phrase candidates ${\cal G}({\cal C}_t)$ w.r.t. the
restricted corpus $\mathcal{C}_t$ only contains those strings that occur in different left and right contexts within $\mathcal{C}_t$. 
%
Now, when processing sentence $S_{t+1}$ we compute ``island'' substrings $V$ characterized by the following property:
\begin{enumerate}
\item $V$ is a maximal infix of $S_{t+1}$ which does not contain as a substring any phrase candidate  in $\cal{G}({\cal C}_t)$ bounded with function words, $\mathcal{F}(\mathcal{P})$.
\end{enumerate}
``Island substrings'' $V$ with these properties are extended on both sides until we reach the sentence border or a delimiting function word. 
Extended island strings $\hat V$ are treated as candidates for phrases.
For example, if sentence $S=S_{t+1}$ has the form

\texttt{The\_Commission\_proposal\_is\_a\_start\_but\_it\_is\_not\_enough.}

\noindent it might be the case ``\texttt{The\_}'', ``\texttt{\_is\_a\_}'', ``\texttt{\_but\_}'', ``\texttt{\_it\_is\_}'', and ``\texttt{\_not\_}'' are already elements
of ${\cal G}({\cal C})_t$, but none of the remaining words in the sentence has this property. 
The strings ``\texttt{Commission\_proposal}'', ``\texttt{start}'' and ``\texttt{enough.}'' are islands not covered by these phrase candidates: 

\texttt{\underline {The\_}Commission\_proposal\underline{\_is\_a\_}start\underline{\_but\_it\_is\_not\_}enough.}

\noindent Assume that when decomposing the sentence these strings are
decomposed in a brute force way into substrings like ``\texttt{C}'', ``\texttt{ommission}'', ``\texttt{pro}'', ``\texttt{po}'', ``\texttt{osal}''
and so on. None of these smaller strings is bounded by function words on both sides. In this situation we guess that

\texttt{\_Commission\_proposal\_},\quad \texttt{\_start\_}\quad and\quad \texttt{\_enough.} 

\noindent are phrases. Applying this simple procedure we extend the set of phrases, $\mathcal{P}(\mathcal{C})$ with these
extended islands. Results obtained are illustrated in Table~\ref{table_simple_extensions_en}.

\begin{table}
\begin{center}
{\small
\begin{tabular}{|c|}\hline
\texttt{\_implemented\_without\_delay\_}\\\hline
\texttt{\_inaugural\_speech\_}\\\hline
\texttt{\_wilful\_murder\_}\\\hline
\texttt{\_medium-sized\_businesses\_}\\\hline
\texttt{\_Echternach\_procession\_}\\\hline
\texttt{\_asset\_redistribution\_}\\\hline
\texttt{\_displaying\_intolerable\_inertia\_}\\\hline
\texttt{\_dogged\_insistence\_}\\\hline
\texttt{\_social\_environment\_}\\\hline
\texttt{\_workable\_solution\_}\\\hline
\texttt{\_bears\_impressive\_witness\_}\\\hline
\texttt{\_non-consultant\_hospital\_doctors\_}\\\hline
\texttt{\_road\_accidents\_}\\\hline
\texttt{\_modern-day\_slaves\_}\\\hline
\texttt{\_unfair\_advantages\_}\\\hline

\hline
\texttt{\_tatenlos\_zusehen\_}\\\hline
\texttt{\_bromierte\_Flammschutzmitteln\_}\\\hline
\texttt{\_mehrjährige\_Ausrichtungsprogramm\_}\\\hline
\texttt{\_gewisse\_Entwicklungsrückstände\_aufgeholt\_}\\\hline
\texttt{\_gutgemeinte\_Ratschläge\_}\\\hline
\texttt{\_falschen\_Illusionen\_wecken\_}\\\hline
\texttt{\_angestrebten\_Programmplanungsziele\_}\\\hline
\texttt{\_Reformmaßnahmen\_eingeleitet\_}\\\hline
\texttt{\_geleisteten\_Wirtschaftshilfe\_}\\\hline
\texttt{\_hemmungslosen\_Subventionswettlauf\_}\\\hline
\texttt{\_funktionierenden\_Wettbewerbsordnung\_}\\\hline
\texttt{\_monatelange\_Renovierung\_}\\\hline
\texttt{\_rotgrüne\_Bundesregierung\_}\\\hline
\texttt{\_Wasserverschmutzung\_ankämpfen\_}\\\hline
\texttt{\_aussichtsreichsten\_Anwärter\_}\\\hline
\texttt{\_Einzelfall\_herauszugreifen\_}\\\hline
\texttt{\_Strukturfondsmittel\_eingehen\_}\\\hline
\texttt{\_zynische\_Doppelzüngigkeit\_}\\\hline

\end{tabular}
}
\end{center}
\caption{``Extended island phrases'' extracted from an excerpt of the English and German EU Parliament Corpus.}\label{table_simple_extensions_en}
\end{table}

{\bf 2. Instances of island schemes.\ } In a second step we extended this idea. For each sentence $S_t$
in the original corpus a copy $S_t^a$ was introduced where the longest infixes which cannot be covered by phrases in $\mathcal{P}_t$
are substituted by a new character ``UNK''. For example, the above sentence would be transformed to

\texttt{The\_UNK\_is\_a\_UNK\_but\_it\_is\_not\_UNK}

\noindent In this way we obtain a new corpus $\mathcal{C}^a=\{S_n^a\}_{n=1}^N$ consisting of abstracted sentence transformations of the original corpus.
Abstract phrases detected in this corpus can be pulled back to $\mathcal{C}$, in this way recognising patterns that we were
unable to detect originally. For example, if \texttt{The\_UNK\_is\_a\_UNK\_} is a phrase in $\mathcal{C}^a$ we may conclude that \texttt{The\_Commission\_proposal\_is\_a\_start\_} could be treated also as a phrase in the original corpus, $\mathcal{C}$.
For phrase detection in $\mathcal{C}^a$ we apply the algorithm described in Section~\ref{Sec2Sub1}. Afterwards we can simply take the
resulting phrases containing UNK in $\mathcal{P}^a$ - called {\em island schemes} - and replace the special symbol by the original island.
Table~\ref{table_complex_extensions_en} 
represents some island schemes and instances obtained by applying this procedure to the excerpts of English and German EU Parliament Corpus.

\begin{table}
{\small
\begin{tabular}{|c|c|}
\hline
Sample phrase in $\mathcal{C}^a$ & Reconstructed phrase in $\mathcal{C}$ \\ \hline
\texttt{\_of\_UNK\_and\_the\_UNK\_of\_UNK\_} & \texttt{\_of\_liability\_and\_the\_payment\_of\_compensation\_} \\\hline
\texttt{\_of\_UNK\_and\_the\_UNK\_of\_UNK\_}&\texttt{\_of\_detention\_and\_the}\\&\texttt{\_criminalisation\_of\_asylum-seekers\_} \\\hline
\texttt{\_of\_UNK\_the\_trust\_of\_} &\texttt{\_of\_eroding\_the\_trust\_of\_}\\\hline
\texttt{\_of\_UNK\_the\_trust\_of\_} &\texttt{\_of\_regaining\_the\_trust\_of\_}\\\hline
\texttt{\_UNK\_to\_the\_UNK\_of\_the\_} &\texttt{\_pay\_particular\_attention\_to\_the\_wording\_of\_the\_}\\\hline
\texttt{\_UNK\_to\_the\_UNK\_of\_the\_} &\texttt{\_looks\_forward\_to\_the\_enactment\_of\_the\_}\\\hline
\texttt{\_UNK\_to\_the\_UNK\_of\_the\_} &\texttt{\_victim\_to\_the\_rationalisations\_of\_the\_}\\\hline
\texttt{\_UNK\_in\_terms\_of\_UNK\_}&\texttt{\_measured\_in\_terms\_of\_gross\_domestic\_product\_}\\\hline
\texttt{\_UNK\_in\_terms\_of\_UNK\_}&\texttt{\_extensive\_in\_terms\_of\_compliance\_}\\\hline
\texttt{\_UNK\_in\_terms\_of\_UNK\_}&\texttt{\_benefits\_in\_terms\_of\_welfare\_}\\\hline
\texttt{\_UNK\_of\_UNK\_which\_}&\texttt{\_phenomenon\_of\_solidarity\_which\_}\\\hline
\texttt{\_UNK\_of\_UNK\_which\_}&\texttt{\_canons\_of\_beauty\_which\_}\\\hline
\texttt{\_UNK\_of\_UNK\_which\_}&\texttt{\_rock\_of\_Sisyphus\_which\_}\\\hline

\hline
\texttt{\_UNK\_zur\_UNK\_der\_UNK} & \texttt{\_integrierten\_Strategien\_zur}\\&\texttt{\_Wiederbelebung\_der\_Beziehungen\_} \\\hline
\texttt{\_UNK\_zur\_UNK\_der\_UNK}&\texttt{\_Zuverlässigkeitserklärung\_zur\_Ordnungsmäßigkeit}\\&\texttt{\_der\_zugrundeliegenden\_Vorgänge\_} \\\hline
\texttt{\_in\_den\_UNK\_und\_UNK\_} &\texttt{\_in\_den\_Medien-\_und\_Kulturnetzen\_}\\\hline
\texttt{\_in\_den\_UNK\_und\_UNK\_} &\texttt{\_in\_den\_Entbindungsheimen\_und\_Krankenhäusern\_}\\\hline
\texttt{\_UNK\_des\_UNK\_der\_Union\_} &\texttt{\_Konsolidierung\_des\_Wirkens\_der\_Union\_}\\\hline
\texttt{\_UNK\_des\_UNK\_der\_Union\_}&\texttt{\_Identitätsfaktor\_des\_Bildes\_der\_Union}\\\hline
\texttt{\_UNK\_des\_europäischen\_UNK\_}&\texttt{\_Büro\_des\_europäischen\_Bürgerbeauftragten}\\\hline
\texttt{\_UNK\_des\_europäischen\_UNK\_}&\texttt{\_Teilung\_des\_europäischen\_Kontinents\_}\\\hline
\texttt{\_UNK\_des\_europäischen\_UNK\_}&\texttt{\_Symbole\_des\_europäischen\_Wissens\_}\\\hline
\texttt{\_UNK\_des\_europäischen\_UNK\_}&\texttt{\_Kernstück\_des\_europäischen\_Landwirtschaftsmodells\_}\\\hline
\texttt{\_in\_UNK\_Weise\_UNK\_}&\texttt{\_in\_lobenswerter\_Weise\_zurückgehalten\_}\\\hline
\texttt{\_in\_UNK\_Weise\_UNK\_}&\texttt{\_in\_bemerkenswerter\_Weise\_gemeistert\_}\\\hline
\texttt{\_in\_UNK\_Weise\_UNK\_}&\texttt{\_in\_unannehmbarer\_Weise\_legitimiert\_}\\\hline
\end{tabular}

}
\caption{Automatically extracted ``island schemes'' and instantiations in excerpts of the English and German EU Parliament Corpora.}\label{table_complex_extensions_en}
\end{table}

\section{Computing subphrase structure and content words}\label{Sec3a}

In this section we further analyse the structure of phrases. As a side result we obtain a method for finding content words (non-function words) of the corpus.  

\subsection{Algorithmic principles}

Now that we have determined a set of function words $\mathcal{F}(\mathcal{C})$ and a multiset of
phrases $\mathcal{P}(\mathcal{C})$ we proceed to structure phrases in a hierarchical way. 
%
%
%
We address this issue in two steps. In the first step we explain how to generate the constituents of
the phrases in $\mathcal{P}(\mathcal{C})$, which we call {\em subphrases}. As a by-product of this step
we will be able to derive a notion corresponding to the common ``words''. Recall that our algorithm just 
analyzes plain sequences of symbols without having any notion of word ``built in''. In a second step, we
continue to structure the resulting subphrases in a hierarchical way, obtaining a form of decomposition tree. 


{\bf 1. Splitting phrases and subphrases.\ } 
Following the discussion in Section~\ref{Sec1} and the model that we developed in Section~\ref{Sec2}, given a phrase $P\in\mathcal{P}(\mathcal{C})$ we look for a decomposition into substrings of $\mathcal{G}(\mathcal{C})$ and determine those
that exhibit the most regular function words. 
Formally, we search for a decomposition of $P$ into a sequence
$P_1,P_2,\dots,P_k$ of general phrase candidates $P_i$ such that the sequence of induced function words $\overline{F}=(F_1,F_2,\dots,F_{k-1})$ optimises the likelihood:
\begin{equation*}
\ell(\overline{F}|\mathcal{P})=\prod_{i=1}^{k-1}\tilde{p}_{pref}(F_i|\mathcal{P})\cdot\tilde{p}_{suf}(F_i|\mathcal{P}).
\end{equation*}
The only difference to the situation above is that we now have fixed the set of function words, requiring that
$F_i\in\mathcal{F}(\mathcal{C})$ for all $1\le i<k$.
%
%
%
Strings $P_i$ only need to be general phrase candidates in $\mathcal{G}(\mathcal{C})$ but may fail to be elements of $\mathcal{P}(\mathcal{C})$. 
In this way, a larger spectrum of meaningful units in the corpus is obtained, beyond the multiset of (full) phrases in $\mathcal{P}(\mathcal{C})$.
 %
We call each $P_i$ a {\bf subphrase} in $\mathcal{C}$. 
Using the same principles again, each subphrase $P$ can be further decomposed into finer subphrases until we arrive at {\bf atomic subphrases} 
that cannot be further partitioned. By $\mathcal{SP}(\mathcal{C})$ we denote the union  of $\mathcal{P}(\mathcal{C})$ with all subphrases obtained from iterated decomposition of phrases and subphrases. 
$\mathcal{SP}(\mathcal{C})$ is called the {\em set of subphrases of the corpus $\mathcal{C}$}. 
The set of atomic subphrases of $\mathcal{C}$ is denoted $\mathcal{ASP}(\mathcal{C})$.

It is easy to see that each subphrase $P$ that is not a sentence prefix (suffix) has a prefix (suffix) which is a function word in $\mathcal{F}(\mathcal{C})$. The {\bf kernel} of $P$ is obtained from $P$ by deleting the prefix (suffix) of $P$ in $\mathcal{F}(\mathcal{C})$ which has the maximal probability $p_{fw}(.\vert \mathcal{P})$ as a function word. If $P$ does not have such a prefix (suffix), we ``delete'' from $P$ the empty prefix (suffix). 
In this way, from the set of phrases $\mathcal{P}(\mathcal{C})$ we obtain the set of {\bf phrase kernels} $\mathcal{P}_K(\mathcal{C})$, from the set of 
subphrases we obtain the set of {\bf subphrase kernels} $\mathcal{SP}_K(\mathcal{C})$, and from the set of atomic subphrases we obtain the 
set of {\bf atomic subphrase kernels} $\mathcal{ASP}_K(\mathcal{C})$. 
It turns out that the atomic subphrase kernels of a phrase $P$ typically are the ``content words'' contained in $P$, i.e., words that are not in the set of $\mathcal{F}(\mathcal{C})$.

{\bf 2. Decomposition trees and functional schemes.\ } 
For a phrase $P$ we may select an optimal decomposition. When we continue to decompose the subphrases obtained, always selecting an optimal decomposition, we 
obtain a {\bf decomposition tree} $t$ for $P$. Each node of $t$ represent a subphrase of $P$. The children of a node 
$P'$ in $t$ are obtained using the set of function words $\mathcal{F}(\mathcal{C})$ and the above optimization principle, selecting one optimal decomposition. A parallel {\bf kernel decomposition tree} $t_K$ is obtained replacing each node in $t$ by its kernel. Using the two trees and looking at the leaves, each phrase $P$ has a representation as a sequence of function words and atomic kernels. For phrases $P$ that do not correspond to a sentence prefix or suffix, $P$ is described as  
$$P= F_0 A_1 F_1\ldots F_{n-1} A_n F_n$$
where $F_i \in \mathcal{F}(\mathcal{C})$ and the strings $A_i$ represent the leaves of $t_K$.  In this situation, the sequence 
$F_0 \vert F_1 \vert \ldots \vert F_n$ is called a {\bf functional scheme} for $P$. For sentence prefixes (suffixes) function word $F_0$ (resp. $F_n$) has to be omitted. 
A functional scheme can be considered as a rudimentary form of grammar rule. 
In general a (sub)phrase may have several optimal decompositions. Hence, for a given phrase $P$ we obtain a {\bf set of decomposition trees} $T(P)$, 
a parallel {\bf set of kernel decomposition trees} $T_K(P)$, and a {\bf set of functional schemes} $FS(P)$. However, we found that the 
functional scheme for a phrase and the leaf representation $P= F_0 A_1 F_1\ldots F_{n-1} A_n F_n$ is unique in most cases.
From the multiset of phrases of the corpus $\mathcal{C}$ we obtain the 
{\bf multiset $FS(\mathcal{C})$ of all functional schemes} in $\mathcal{C}$, which may be ordered by the number of occurrences.

\subsection{Results for distinct languages and corpora}

\begin{figure}
\includegraphics[width=1.1\textwidth]{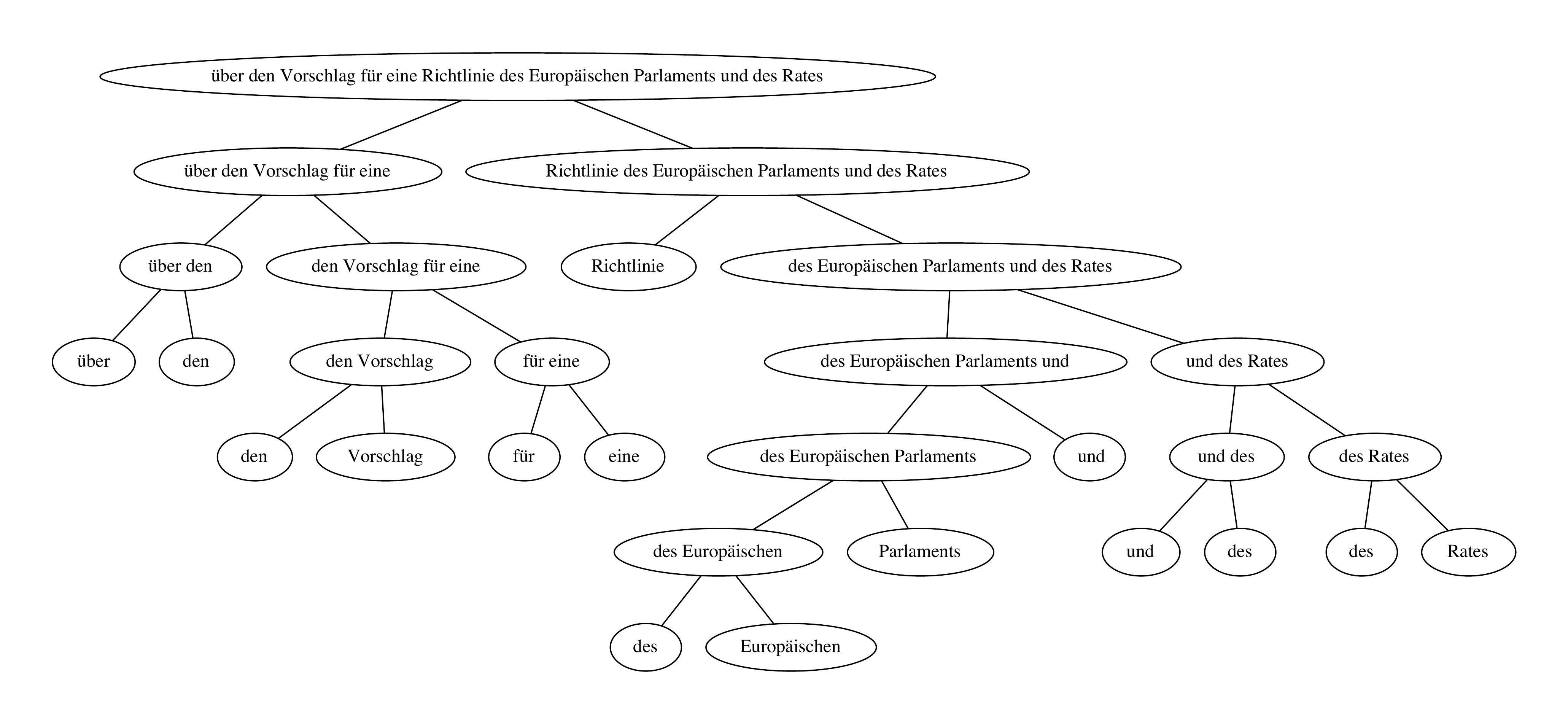}
\includegraphics[width=1.1\textwidth]{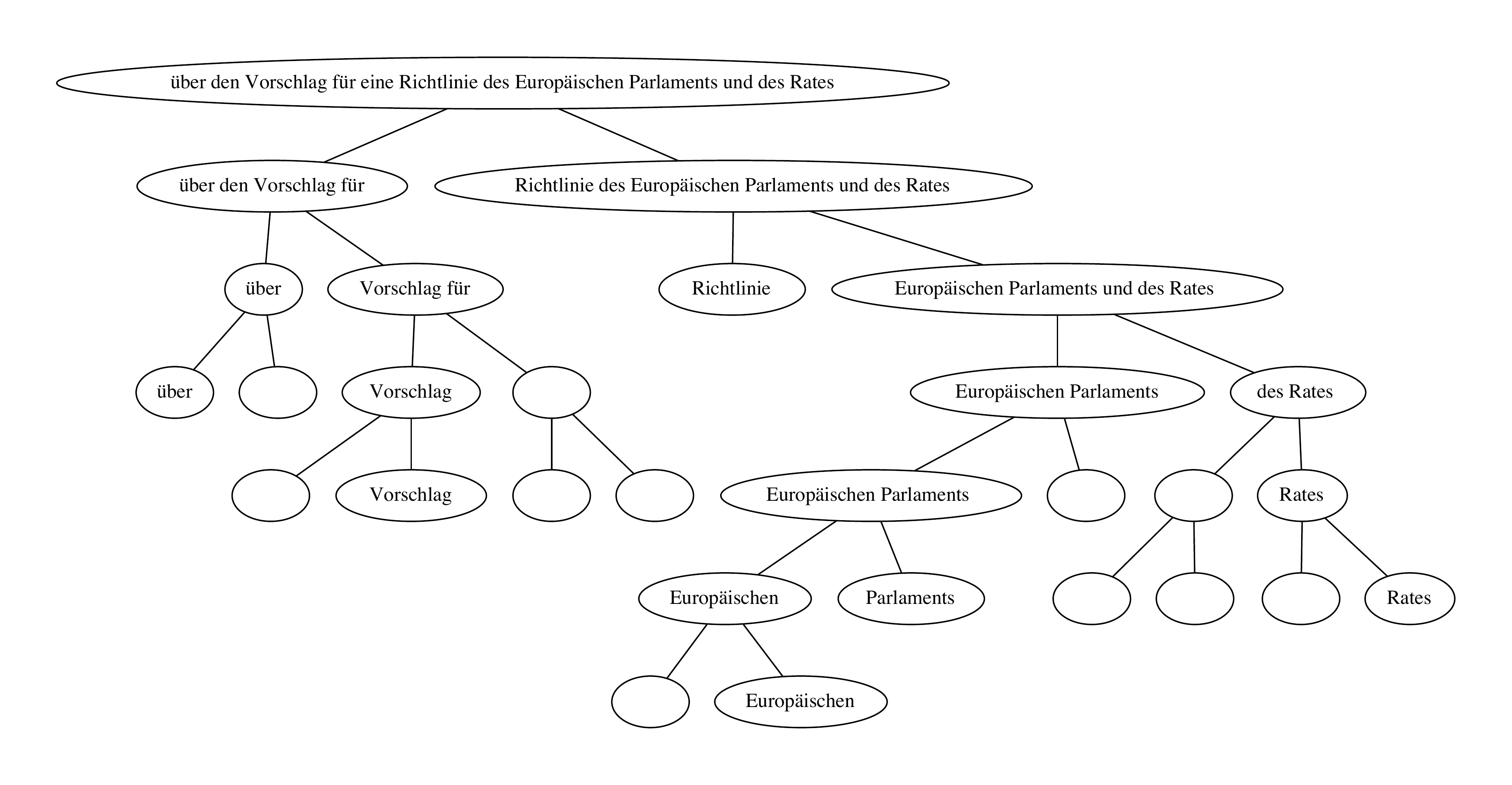}
\caption{German decomposition tree and the parallel kernel decomposition tree.}\label{fig:PhraseTrees}
\end{figure}

As an illustration, in Figure~\ref{fig:PhraseTrees} we present a decomposition tree and the parallel 
kernel decomposition tree for the German phrase
 \begin{quotation}\texttt{\"uber den Vorschlag f\"ur eine Richtlinie des \\ Europ\"aischen Parlaments und des Rates}
\end{quotation}
Empty nodes in the kernel decomposition tree
are caused by phrases that become empty when removing the function word borders. 
The functional scheme for the phrase 
has the form 
$$\vert\texttt{ den }\vert\texttt{ f\"ur eine }\vert \texttt{ des }\vert\vert\texttt{ und des }\vert.
$$
Atomic subphrase kernels obtained  are \texttt{\"uber}, \texttt{Vorschlag}, \texttt{Richtlinie}, \texttt{Euro\-p\"aischen}, \texttt{Parlaments}, and \texttt{Rates}. 

The results for phrase decomposition in European languages are similar. 
As mentioned above, in almost each case, the functional scheme for a phrase turned out to be unique, even if there are several decomposition trees. This shows that the difference between distinct decompositions is mainly an issue of distinct orders in which to decompose (sub)phrases, not an issue of distinct final decomposition parts. 

For all European languages and corpora considered, the set of atomic subphrase kernels obtained is essentially a lexicon of the content words of the corpus. Hence, despite of the fact that we do make any distinction between the distinct symbols in the textual alphabet, words are recognized as the ``atomic'' units from phrase decomposition. 
For Chinese, iterated subphrase splitting often leads to single symbols. Though multi-symbol words in general occur at one level of the decomposition of phrases, we yet do not have good principles to exactly find this point. 

An obvious question is if the functional schemes obtained can be considered as a kind of induced grammar. 
However, there is no direct correspondence between traditional linguistic categories and the spectrum of functional schemes. 
Many of the functional schemes capture small sequence of words of a similar form with many instances, but the distinct instances do not always have a common linguistic functionality.  
In Table~\ref{table:frames_de_en_wiki} 
we show the most frequent functional schemes for the English and German Wikipedia Corpora, respectively. 

On the other hand we also found a considerable number of interesting patterns that point to a special form of 
grammatical construction. 
In Table~\ref{table:frames_de_en_wiki_spec} 
we add  a selection of 
interesting complex schemes with several instances. 

\begin{table}
{\small 
\begin{tabular}{|c|c|c|}\hline
Frame Scheme &\# Occ. & Instance of a phrase with such a frame\\ \hline
 |\, | & 3355652 & \texttt{ show soon }\\ \hline
 the | & 1684261 & \texttt{ the instinct } \\ \hline
 | of & 1011444 & \texttt{ laser of } \\ \hline
. | & 864564 & \texttt{.  Invariant } \\ \hline
| and & 689264 & \texttt{ courier and }\\ \hline
| the & 566956 & \texttt{ exalting the }\\\hline
| to & 551944 & \texttt{ adherence to }\\\hline
| in  & 543862 & \texttt{ dissolves in }\\\hline
of | & 523440 & \texttt{ of civilization.} \\\hline
a |    & 477598 & \texttt{ a tolerance }\\\hline
and | & 474076 & \texttt{ and César }\\ \hline
to | & 408089 & \texttt{ to emigrate }  \\\hline	
the |\,| & 387926 & \texttt{ the approximate area }\\ \hline	
in |   & 339115 & \texttt{ in Brazil, } \\\hline
 the | of       & 301400 & \texttt{ the friends of }\\\hline
 |\,|\,|     & 269382 &\texttt{ future business partner }\\\hline
 | of the        & 268339 & \texttt{ observer of the } \\\hline
  of the |       & 245283 & \texttt{ of the society } \\\hline
 | as    & 206565 & \texttt{ livings as }\\\hline
 | is    & 193940 & \texttt{ reproduction is } \\\hline \hline
%
 |\, | & 2723007 & \texttt{ identische Formen }\\ \hline
 . | & 954597 & \texttt{ . Transaktionen } \\ \hline
 der | & 688544 & \texttt{ der Generaldirektor } \\ \hline
die | & 499024 & \texttt{ die Besatzungen } \\ \hline
| der & 470917 & \texttt{ Gegeninduktionsspannung der }\\ \hline
| und & 566956 & \texttt{ Dardanellen und }\\\hline
, |  & 443004 & \texttt{ , Ausnahmen }\\\hline
| die  & 308887 & \texttt{ beginnen die }\\\hline
und | & 306985 & \texttt{ und Betriebe } \\\hline
|\,|\,|    & 257187 & \texttt{ gute zeitliche Auflösung }\\\hline
| in  & 239913 & \texttt{ Arbeitsgruppen in }\\ \hline
von | & 211493 & \texttt{ Trag\"odie von }  \\\hline	
den | & 201745 & \texttt{ den Willen }\\ \hline	
in |   & 195174 & \texttt{ in Bockenheim } \\\hline
| des       & 193070 & \texttt{ Ordnung des }\\\hline
von |  & 183836 &\texttt{ von Zufallszahlen }\\\hline
des |  &    168460  & \texttt{ des Domes, } \\\hline
im |   & 166382 & \texttt{ im Fr\"uhling } \\\hline
. |\,|     & 131594 & \texttt{ . Daher spielten }\\\hline
  das |  & 130116  & \texttt{ das Mittelschiff } \\\hline
\end{tabular}
}
\caption{Most frequent phrase frames in the English and German Wikipedia Corpus.}\label{table:frames_de_en_wiki}
\end{table}

\begin{table}
{\small 
\begin{tabular}{|c|c|c|}\hline
Frame Scheme & \# Occ. & Instance of a phrase with such a frame\\ \hline
in an | to | & 400 & \texttt{ in an attempt to use }\\ \hline
was | to | the & 400 & \texttt{ was devised to allow the } \\ \hline
a | from the & 399 & \texttt{ a signal from the } \\ \hline
| of | of the & 398 & \texttt{ consisting of members of the } \\ \hline
at its |  & 398 & \texttt{ at its zenith }\\\hline
the | and the |  & 397 & \texttt{ the science and the arts }\\\hline
. The | of a  & 397 & \texttt{. The length of a  }\\\hline
 a | between the    & 396 & \texttt{ a gap between the }\\\hline
| that the | of & 199 & \texttt{ believe that the presence of }\\ \hline
 the | of  other     & 199 & \texttt{ the decisions of other }  \\\hline	
in the | of |\,| & 199  & \texttt{ in the form of compress air }\\ \hline	
 that | only    & 99 & \texttt{ that time only } \\\hline
 the |\,| of | and |      & 99 &\texttt{ the atomic bombings of }\\&&\texttt{ Hiroshima and Nagasaki }\\\hline
the | of | that        & 99 & \texttt{ the branch of philosophy that } \\\hline
| of | games        & 10 & \texttt{ family of card games } \\\hline
| a | system    & 10 & \texttt{ patented a telegraph system }\\\hline
| a  different |    & 10 & \texttt{ take a different approach, } \\\hline
|\,|\,|" is |\,| if and only if  & 5 & \texttt{The field "F" is algebraically }\\ && \texttt{ closed if and only if } \\\hline \hline
%
 mit | \%  & 400 & \texttt{ mit 13,7 \% } \\ \hline
, der | von & 399 & \texttt{ , der Nachfolger von } \\ \hline
| bis zum |     & 396 & \texttt{ reicht bis zum Jahr }\\ \hline
 ein | auf      & 396 & \texttt{ ein Monopol auf }\\\hline
 gibt es in |   & 396  & \texttt{ gibt es in Nordamerika  }\\\hline
über die |\,|    & 395  & \texttt{ \"uber  die Teilung Deutschlands } \\\hline
|\,| als  die    & 395 & \texttt{ wesentlich \"alter als die }\\\hline
 f\"ur |\,| und | & 199 & \texttt{ f\"ur physikalische Chemie }\\&&\texttt{ und Elektrochemie }\\ \hline
 in das | der     & 198 & \texttt{ in das Grundsatzprogramm der }  \\\hline	
als | f\"ur den & 198  & \texttt{ als Ursache f\"ur den }\\ \hline	
| der | am    & 99 & \texttt{ beschloss der Bundesrat am } \\\hline
  f\"ur die | der |     & 99 &\texttt{ f\"ur die Durchsetzung der } \\&&\texttt{ Menschenrechte }\\\hline
nach dem | des |\,|         & 99 & \texttt{ nach dem Ende des NS-Regimes } \\\hline
| der | der Stadt |        & 10 & \texttt{An der Spitze der Stadt stand } \\\hline
| und |\,| Menschen     & 10 & \texttt{ Neandertaler und anatomisch}\\ && \texttt{ modernen Menschen }\\\hline
| als | Kilometer    & 10 & \texttt{ mehr als sechs Kilometer } \\\hline
wurde die | zu einem | der  & 6 & \texttt{Wurde die Stadt zu einem Zentrum der } \\\hline
\end{tabular}
}
\caption{Interesting functional schemes derived from the English and German Wikipedia Corpora.}\label{table:frames_de_en_wiki_spec}
\end{table} 


\section{Applications}\label{Sec4}

In this section we consider a selection of possible applications of the above methods. 
In each case the main point is to show that 
the automated computation of phrases and subphrases of a corpus and the bidirectional structure of the index structure used offer interesting possibilities to approach known problems on new paths. 

\subsection{Corpus lexicology}\label{Sec4aa}

The predominant perspective in lexicology and corpus linguistics is to use 
a repository of corpora for collecting and verifying knowledge about a language or language variety. Under this perspective, language 
is the central object of study, and corpora are just a means to obtain insights about language, e.g., by collecting the vocabulary found in the corpora. There is second perspective, which in general finds less attention: here a given corpus is considered as an independent object of study and the goal is to obtain insights about the ``{\em language of the corpus}''. In this subsection we follow the second perspective. 
The methods described below, as those discussed above, can be applied in an unsupervised and fully automated way. They can be considered as a preliminary step to dynamically 
generate in an ``online'' manner views that characterize the vocabulary and language of a given corpus in distinct ways. 
The long-term objective is to enable ``travels'' in the language of the corpus, to compare the languages of distinct corpora etc. We start with an  study where we try to find the most characteristic vocabulary of a given corpus. Afterwards we look at connections between words, semantic fields and phrase nets and at language style.   

{\bf Characteristic words of a corpus.\ }
The problem of how to find ``important'' terms of a corpus or document collection has found considerable attention (see Section~\ref{Sec5}). Since both very rare and very frequent terms in general are inappropriate, there is no simple solution. Here we suggest a method that takes the composition of 
sentences and phrases into account.
In order to find characteristic words of the corpus $\mathcal{C}$ we first try to find 
the most ``characteristic'' or ``specific'' child $P_i$ of a phrase $P\in \mathcal{P}(\mathcal{C})$ 
in the decomposition trees $t\in T(P)$. As we see below, at this step ``rare'' language units are prefered. Afterwards, looking at those kernels in $\mathcal{ASP}_K(\mathcal{C})$ that appear as the kernels of many ``most characteristic subphrases'', we derive a ranked set of ``characteristic words'' or ``key words'' of the corpus $\mathcal{C}$. At this step, terms that are too rare obtain a low score.  
In order to formalise the intuitive notion of ``characteristic", we argue probabilistically. The intuition is that we want to decrease the uncertainty to obtain a longer phrase $P$ starting from a shorter subphrase $P_i$. Probabilistically, this means 
that we need to maximise the conditional probability $p(P|P_i)$ where $P_i$ ranges over all subphrases of $P$. Now, clearly:
\begin{equation*}
p(P|P_i)=\frac{p(P,P_i)}{p(P_i)}
\end{equation*}
and if we look at the probability for these events as the event that the particular strings \emph{occur}, then it
is obvious that $p(P,P_i)=p(P)$ because $P_i$ is a substring of $P$. Thus, 
if we naively consider the probability $p$ as empirical probability, then the probability $p(P|P_i)$ is simply
the ratio of the occurrences of $P_i$ that imply an occurrence of $P$, i.e.:
\begin{equation*}
p(P|P_i)=\frac{occ(P)}{occ(P_i)}.
\end{equation*}
Practically, the maximal value is simply obtained by selecting
the child $P_i$ of $P$ with the smallest number of occurrences $occ(P_i)$ in the corpus.
In our experiments, if two children have the same number of occurrences we simply selected the leftmost child. 

Using the above measure for specificity we assign to each subphrase $P \in \mathcal{SP}(\mathcal{C})$ its most 
specific child $P'$. We write $P'= \parent(P)$. Since a given subphrase $P'$ in general acts as the most specific child of 
several superphrases $P$ we obtain a forest 
$$\mathcal{MSS}_\mathcal{C} = (\mathcal{SP}(\mathcal{C}),\parent)$$
called the {\bf forest of most specific subphrases}. 
The roots of $\mathcal{MSS}_\mathcal{C}$ are the ``characteristic'' elements in the set of atomic subphrases, 
$\mathcal{ASP}(\mathcal{C})$. The nodes of $\mathcal{MSS}_\mathcal{C}$ are the elements of $\mathcal{SP}(\mathcal{C})$, leaves representing maximal phrases.
As before, as an alternative view we may look at the modified forest 
$\mathcal{MSS}^K_\mathcal{C}$ where we replace each subphase by its kernel.  
The roots of this forest, which represent ``characteristic'' content words, 
are called {\bf characteristic atomic kernels}.

As a measure for the importance of an atomic phrase $P$ in $\mathcal{ASP}(\mathcal{C})$ we may use the number $\rho_{\cal C}(P)$ of distinct phrases in $\mathcal{P}(\mathcal{C})$ that are descendants of $P$ in the forest $\mathcal{MSS}_\mathcal{C}$. 
In the experiments described below we found that selecting atomic phrases (or their kernels) with high $\rho_{\cal C}$-score in fact yields a characteristic profile for a given corpus.

In our experiments we computed a ranked list of characteristic atomic kernels for our fragment of the Medline corpus, following the above method. The 
60 top-ranked kernels are shown in Table~\ref{table:Cap2}.  For the sake of comparison, similar ranking list were computed for the English Europarl corpus and for $2\%$ of the English Wikipedia. 
In Table~\ref{table:Cap2}, pairs (n/m) present Europarl ranks (n) and Wikipedia ranks (m). Entries ``-'' mean that the entry did not occur in the ranked list for the corpus.   
\begin{table}[h]
{\small
\begin{tabular}{l l l  }
1. observed (2310/602) &21. one (17/3) &41. tumor (-/10737)\\
2. those (121/30) &22. similar (1341/213) &42. children (1001/177)\\
3. used (220/1) &23. performed (5841/192) &43. identified (1864/608)\\
4. only (9/5) &24. control (362/150) &44. use (151/6)\\
5. treatment (2223/1106) &25. lower (3088/550) &45. different (231/194)\\
6. groups (361/142) &26. demonstrated (1167/1900) &46. examined (4705/5826)\\
7. present (127/114) &27. disease (3185/1674) &47. normal (3079/1211)\\
8. obtained (3743/1141) &28. age (1224/896) &48. time (41/2)\\
9. increase (534/358) &29. cells (-/1123) &49. blood (-/1376)\\
10. increased (991/386) &30. data (1947/277) &50. cases (561/316)\\
11. reported (4701/340) &31. reduced (1340/773) &51. studied (4507/846)\\
12. found (719/8) &32. decreased (5109/3881) &52. specific (284/462)\\
13. results (723/432) &33. risk (556/1182) &53. subjects (1068/2110)\\
14. higher (1253/505) &34. changes (728/270) &54. decrease (3735/2992)\\
15. patients (4104/1897) &35. group (198/66) &55. protein (7516/1941)\\
16. significantly (6063/958) &36. previously (3324/676) &56. 3 (720/779)\\
17. patient (-/3414) &37. developed (633/26) &57. compared (2692/1199)\\
18. cancer (-/2935) &38. high (532/226) &58. women (276/199)\\
19. study (1271/134) &39. effective (416/522) &59. investigated (3658/5008)\\
20. expression (1333/1666) &40. factors (1920/1089) &60. evaluated (8548/7615)\\
\end{tabular}

}
\caption{Ranked list of characteristic atomic kernels from Medline, parallel ranks in the Europarl corpus (EP) and in the Wikipedia (W).}\label{table:Cap2}
\end{table}
The ranked lists for the corpora, despite of the large variety of word types in the lists, are radically distinct. When taking the top 60 
kernels from Medline, only three (``only'', ``one'', ``time'') occur among the top 60 words in the ranking list for the Europarl corpus, and only seven (``used'', ``only'', ``found'', ``one'', ``developed'', ``use'', ``time'') occur among the top 60 words in the ranking list for the Wikipedia corpus. 
This shows that top segments of the ranking lists yield an interesting ``profile'' for a given corpus, which might also be useful for classification tasks and for stylistic studies (see below).

\begin{figure}
\begin{center}
\includegraphics[width=0.8\textwidth]{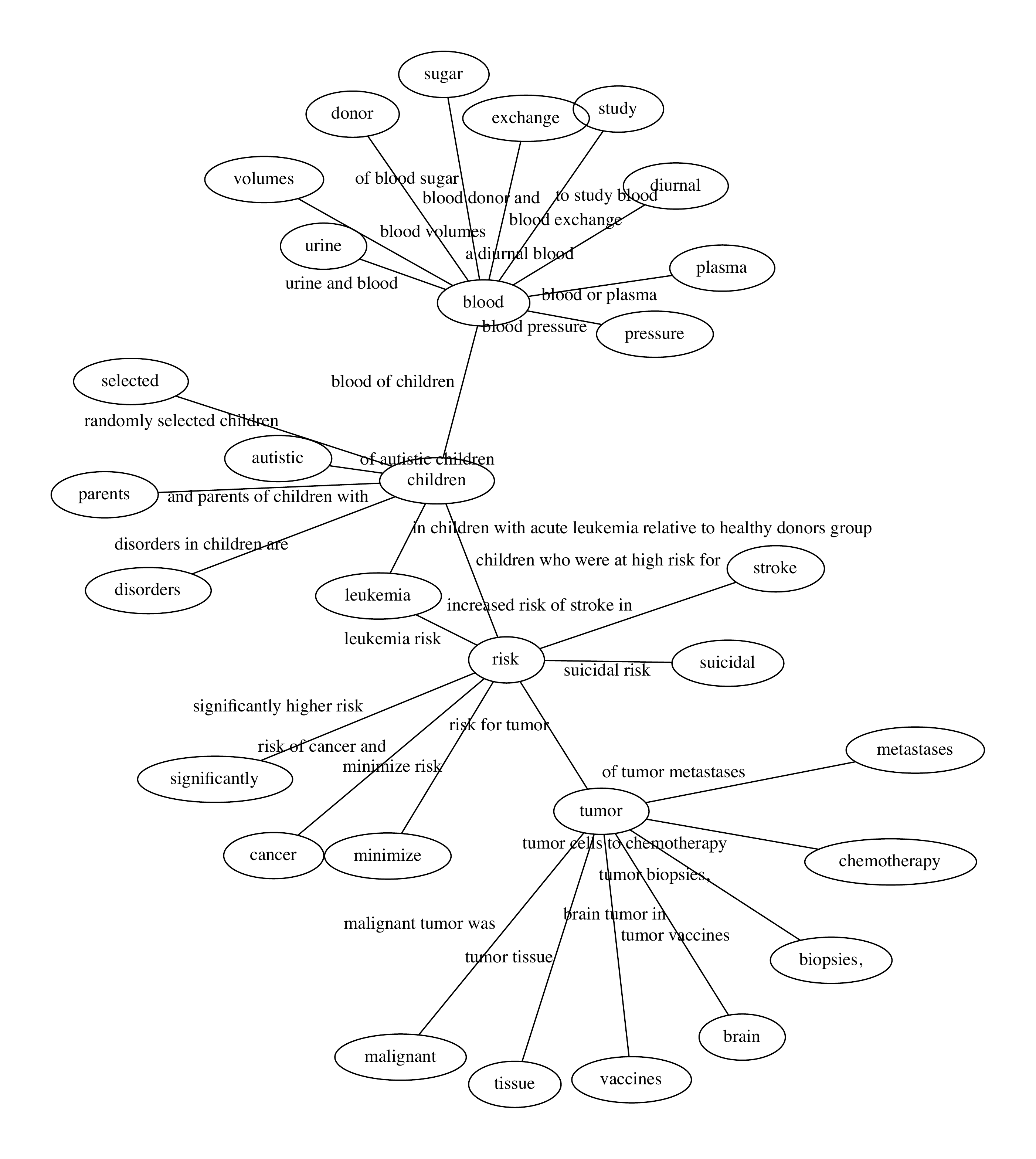}
\end{center}
\caption{Network derived from Medline corpus using co-occurrences in phrases.}\label{fig:Networks1}
\end{figure}
\begin{figure}
\begin{center}
\includegraphics[width=0.8\textwidth]{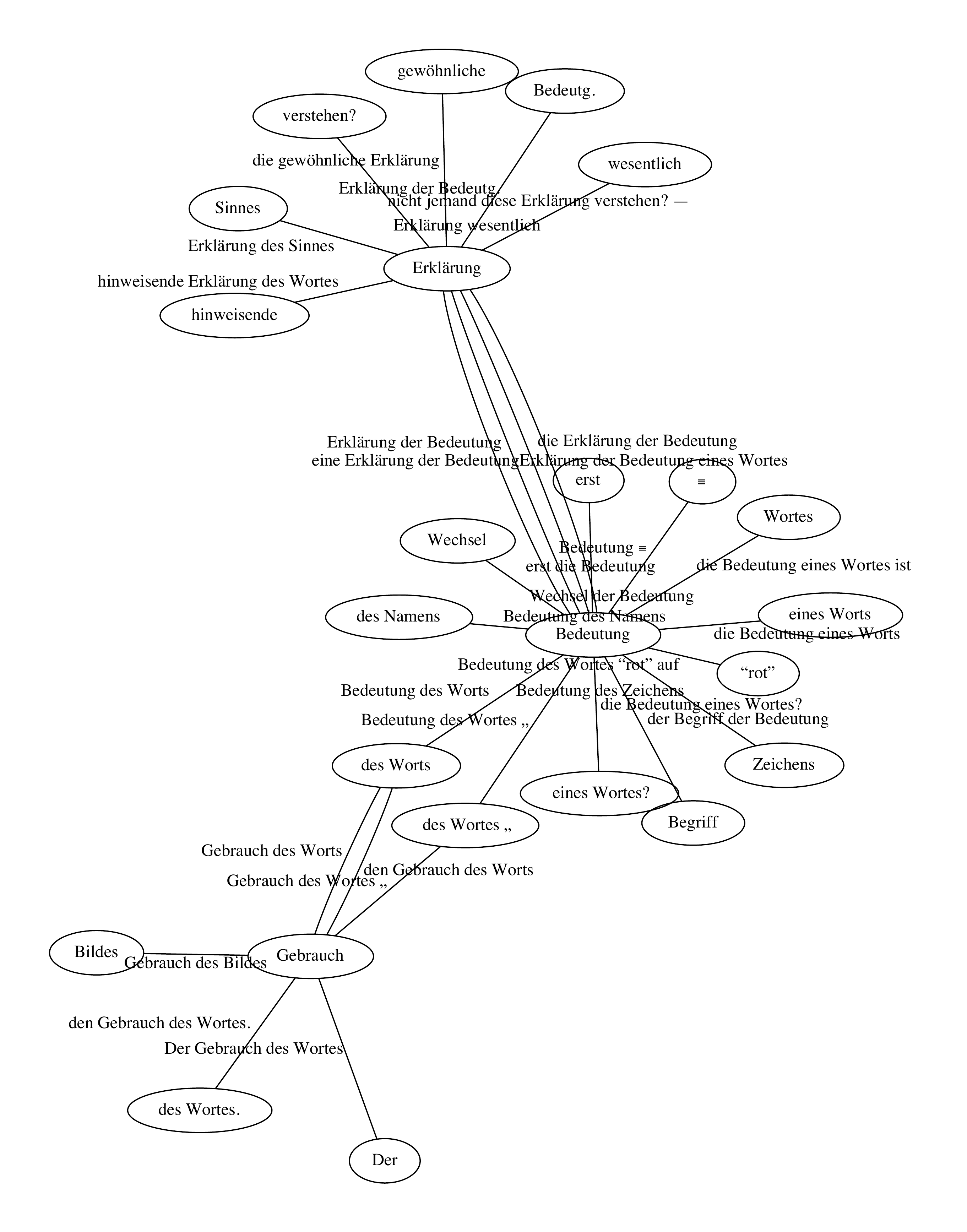}
\end{center}
\caption{Network derived from Wittgenstein corpus using co-occurrences in phrases. }\label{fig:Networks3}
\end{figure}

{\bf Connections between words, semantic fields and phrase nets.\ }
Automated approaches for exploring the paradigmatics of lexical units typically look at the co-occurrence of terms in documents, paragraphs, sentences or fixed size neighbourhoods \cite{SchuetzePedersen93,storjohann2010lexical}. Another, more ``syntactic'' view is obtained when looking at the co-occurrence of terms in {\em phrases}. To this end, 
the forest of most specific subphrases $\mathcal{MSS}_\mathcal{C}$ can be used. We compute the network
$G=(\mathcal{ASP}(\mathcal{C}), E)$ where $E$ is the set of all edges of the form $e=(A_1,P,A_2)$ where $P\in\mathcal{P}(\mathcal{C})$ 
is a descendant of the root nodes $A_i$ in $\mathcal{MSS}_\mathcal{C}$ ($i=1,2$). In the graph $G$, two atomic phrases $A_1$ and $A_2$ are strongly interlinked if they co-occur in many full phrases. 
Note that phrases $P$ can have variable length and we do not ask if $A_1$ and $A_2$ are direct neighbours in $P$. 
%
Figures~\ref{fig:Networks1}
and~\ref{fig:Networks3} 
show some of these graphs. Networks of this kind, when visualized in an appropriate way, offer interesting possibilities for mining and exploring corpora and related corpus vocabulary. 
Figure~\ref{fig:Networks1} starts from four characteristic words from Table~\ref{table:Cap2}, ``blood'', ``children'', ``risk'', and ``tumor''. Presented are atomic phrases that co-occur with these words in phrases, and the phrases where the co-occurrence appeared. Figure~\ref{fig:Networks3} is obtained in a similar way from the Wittgenstein corpus, starting from ``Erkl\"arung'', ``Bedeutung'' and ``Gebrauch''. 



{\bf Mining language style.\ }
It is a well-known phenomenon that the style even of {\em official} language differs when looking at distinct communities, fields or domains. When entering a new community, even native speakers often have difficulties to ``adapt'' language style and to meet the typical phrases and constructions of the domain. For foreign language learners this task is even more complex. 
Many phrases automatically extracted from corpora using the above methods reflect the typical style of the corpus domain.
For mining ``stylistic'' phrases we may start from subphrases such as  `` we '', `` my '', and ``it is '' and then look at the full phrases containing these strings. These phrases exhibit another specific ``face'' for each corpus. 
Below we present phrases including the strings 
`` my '', and ``it is '' derived (a) from the Europarl corpus, and (b) from Medline. 
It is important to note that the phrases extracted may contain the given string at any position, such as in ``as a conclusion {\em it is} suggested that'', or in ``express {\em my} regret at the''. 

{\small 

{\em  ``my'' in Europarl:\ } my wholehearted support - my compliments to - extend my special thanks to - my congratulations to - my colleague , Commissioner Wallstr\"om - In my opinion , this is the only way - In my view , this - close to my heart - to thank my colleagues on the committee for - I expressed my views on - It is my belief that - On behalf of my - my honour to present to you - my group welcomes this - express my regret at the ...

{\em  ``my'' in Medline:\ } to my knowledge this is the first - my own experience - my personal experience - studies in my laboratory - my professional life - of my research career - my assessment - my findings suggest that - answer to my question - my colleagues and i have developed ... .
\vskip 1mm

{\em  ``it is'' in Europarl:\ } it is extremely important - it is therefore necessary to - it is a question of  - it is , however , - it is in this spirit that - it is especially important - 
it is high time that - it is indeed the case that - it is my belief that - it is worth pointing out - it is unacceptable that - it is not enough to ...\\
{\em ``it is'' in Medline:\ } from these results it is concluded that - it is suggested that - it is shown that - it is argued that - it is postulated that - however it is not known whether - therefore it is suggested that - it is tempting to speculate that - indicating that it is - it is important to understand - furthermore it is shown that - as a conclusion it is suggested that ...
}


\subsection{Terminology extraction}\label{SubSec4.1}

Automated terminology extraction helps to build up terminological dictionaries and thesauri, it has often been studied in the literature (Section~\ref{Sec5}). 
Most methods suggested use characteristic sequences of POS, word association measures such as log likelihood, mutual information etc., and corpus comparison to find terminological expressions in a given corpus. 
Here we suggest a three step approach only based on the techniques described above and using distinct corpora. 
In Step 1 we compute the characteristic atomic kernels of phrases for the given corpus, only relying on the input corpus. 
In Step 2, given the set of characteristic atomic kernels we try to single out ``terminological'' kernels  (e.g., ``cancer'') using the ranked lists of characteristic atomic kernels obtained from another corpora for comparison.  
In Step 3, given a terminological kernel (``cancer'') we try to find terminological (multi-word) expressions that contain this kernel as substring (``female lung cancer'', ``prostate cancer''). 

{\bf Step 1 - characteristic atomic kernels.\ } Our starting point for searching kernels is the decomposition of phrases described in Section~\ref{Sec3a}. We compute the ranked list of characteristic atomic kernels. 
In general there are much more truly ``characteristic'' kernels than ``terminological'' kernels. Even if there is hardly a general agreement of what ``terminological'' exactly means, typically, mainly noun phrases are considered as terminological expressions. The list in  Table~\ref{table:Cap2}
show that words of many distinct types can be ``characteristic'' for a given corpus. 


 {\bf Step 2 - terminological kernels.\ } When trying to filter out ``terminological'' kernels from the list of ``characteristic'' kernels we could of course use POS or other type of linguistic information. However, for the sake of consistency of the approach we do not use external or human linguistic knowledge here. The numbers found in  Table~\ref{table:Cap2} suggest that ``terminological'' kernels typically do not appear in the top segment of the list obtained for the Wikipedia corpus. If we ignore kernels with a rank below 1,000 in one of the two corpora used for comparison, from the Medline top 60 segment we obtain the following list of kernels 
 \begin{quotation}
 {\tiny treatment (2223/1106), obtained (3743/1141), patients (4104/1897), patient (-/3414), cancer (-/2935), expression (1333/1666), demonstrated (1167/1900), disease (3185/1674), cells (-/1123), decreased (5109/3881), factors (1920/1089), tumor (-/10737), examined (4705/5826), normal (3079/1211), blood (-/1376), subjects (1068/2110), decrease (3735/2992), protein (7516/1941), compared (2692/1199), investigated (3658/5008), valuated (8548/7615)}
 \end{quotation}
We still find entries that are usually not considered as ``terminological'', but clearly a much better focus is obtained. With larger bounds, the percentage of non-termino\-logical kernels can be further reduced, a suitable compromise between precision and recall needs to be defined for each application.

{\bf Step 3 - kernel expansion.\ } 
For each of the remaining top-ranked phrase kernels we consider the list of all phrases that contain this kernel as a substring. In this way, kernels are expanded to the left, to the right or in both directions. 
Since our phrases typically come with function word delimiters we clean each set by eliminating left and right ``borders'' that appear in our list of function words (see above). 
Cleaned substrings are ranked by frequency of occurrence. As a result we obtain a set of multi-word expressions. 
The characteristic features of the terminological expression found in this way are the following
\begin{enumerate}
\item the {\em number of words} in these expressions is {\em not fixed},
\item the {\em kernels} used may {\em appear at any position} in such a multi-word expression. 
\end{enumerate}
As an illustration, in Appendix A we present upper parts of the ranked lists obtained for the kernels ``disease'', ``syndrome'', and ``inflammatory''. 
In most cases, ``disease'' occupies the last position of the phrase. Expressions can be long - some examples from the full list are ``risk factor for cerebrovascular disease'', or ``intestinal failure-associated liver disease''. Examples where ``disease'' is not at the last position are ``disease-related modification of exposure'' and ``disease is known to be caused''. For ``syndrom'', an example is 
``metabolic syndrome components''. 
In contrast, ``inflammatory'' typically is found at the beginning of an expression, as expected. Examples not following this rule are ``proinflammatory cytokine'' and ``nonsteroidal anti-inflammatory drugs''. We did do not use any kind of additional cosmetics (which would easily be possible in practice). From our point, this third step is working extremely well.

\subsection{Automated query expansion for search engines and free context size answer browsing}\label{SubSec4.2}

Querying a large corpus can be a difficult task since it is not simple to estimate in advance the number of answers to a search query. If the query is too specific, the number of answers might be small or empty, in other cases the answer set becomes extremely large and a manual inspection is not possible. In order to support users, search engines in the Internet often present a selection of possible right expansions to a unspecific string. For example, when typing 
``John'' in the search window of the English Wikipedia, some expansions such as ``John McCain'', ``John Lennon'', and others (s.b.) are suggested. 

The index-based technology presented above can be used to automatically produce interesting expansions.
A given short query is just treated as a kernel in the sense of Section~\ref{SubSec4.1}. 
Just using Step~3 ``kernel expansion'' of the above procedure, interesting {\em right and left} expansions for a query are found. In order to judge how well this approach works we indexed the complete English Wikipedia\footnote{The corpus, Wikipedia dumps May 2014, used for these experiments was split into sentences by PUNKT,~\cite{kiss2006}, and afterwards processed by our approach from Section~\ref{Sec2Sub1}.}
lists of possible expansions for a number of short queries. Possible expansions were ranked by the number of occurrences. Topmost Expansion suggestions for ``John'', ``Peter'', ``Blue'' and ``Brown'' are presented in Appendix B. 
We then compared this with Wikipedia's expansion suggestions. Main insights obtained are:
\begin{enumerate}
\item 
Wikipedia only suggests a small number of expansions. The list of possible expansions produced by our method is much larger. 
\item
Wikipedia only produces right expansions for a query. Our lists also include left or two-sided expansion suggestions.
\item
As to recall, most of Wikipedia's suggestions are also found in our lists. In most exceptional cases, variants of Wikipedia's suggestions are found. 
\end{enumerate}
The third point is illustrated in Table~\ref{tableSmaller} where we show the full list of Wikipedia's expansion suggestions (June 18, 2015) for ``John'', ``Peter'', ``Blue'', ``Brown'', ``Italian'', and ``French''. We show which suggestions are also found using our index-based ``kernel expansion'' method, and where we found variants of Wikipedia's suggestions. The recall is excellent.  
In many cases, distinct variant phrases of Wikipedia's suggestions were found in addition. Some of these are mentioned in Table~\ref{tableSmaller}.
\begin{table}
{
\begin{center}
{\tiny
\begin{tabular}{|l|l|}\hline
John McCain   &  yes  (356)  \\ \hline
John Lennon    &  yes  (602)\\ \hline
John, King of England &  yes (9)   \\ \hline
John, Paul, George, and Ringo &  yes (2), also variants found   \\ \hline
John Tolkein &  ``J. R. R. Tolkien'' found for kernel ``J. '' \\ \hline
John A. Mcdonald    &  ``John A. Macdonald'' found (62) \\ \hline
John Wayne Gacy    &  yes (27) \\ \hline
John Roberts    &  yes (140)\\ \hline
Peter Ilyich Tchaikovsky & yes (3) ``Pyotr Ilyich Tchaikovsky'' found for kernel ``Pyotr'' \\ \hline
Peterborough    &  yes (1098)\\ \hline
Peter Parker, the Spectacular Spider-Man    & yes (8) \\ \hline
Peter Sellers    &  yes (187)\\ \hline
Peter the Great    &  yes (311)\\ \hline
Peter Jennings    &  yes (47) \\ \hline
Peter IV of Portugal &  yes (3)   \\ \hline
Peterloo Massacre    &  yes (5) \\ \hline
Blue    &  yes  (3251)\\ \hline
Blue's Clues    &  yes (55) \\ \hline
Blue whale    &  yes (19) \\ \hline
Bluetooth    &  yes (358)\\ \hline
Blues    &  yes (1931)\\ \hline
Blue ray disc    &  no \\ \hline
Blue-green algae    &  yes (2) \\ \hline
Blue iguana    &  ``Blue iguana,'' found (3) \\ \hline
Blues rock    &  yes (4) \\ \hline
Blue1    &  yes (8) \\ \hline
Brown    &  yes (6564)\\ \hline
Brown v. Board of Education    &  yes (32)\\ \hline
Brown Mackie College    & yes (7) \\ \hline
Brownhills     &  yes (49) \\ \hline
Brown Dog affair    &  yes (3) \\ \hline
Browning Hi-Power    &  yes (11)\\ \hline
Brown  gold  &  no \\ \hline
Brown dwarf   &  ``brown dwarf'' (84) found for kernel ``brown'' \\ \hline
Brown-grey  &  ``brown-grey'' grey( 37)found for kernel ``brown'' \\ \hline
Browning M2 &  yes (5) \\ \hline
Italian &   yes (5409)\\ \hline
Italianate architecture & yes (14)\\ \hline
Italians &   yes (1137)\\ \hline
Italian language & yes (166)\\ \hline
Italian War of 1521-26 & ``the Italian War of 1521-1526,'' found (4) \\ \hline
Italian films of the 2010s & ``Italian films'' found \\ \hline
Italian ballet & yes (3) \\ \hline
Italian War of 1542-46 & ``the Italian War of 1542-46'' found (2)\\ \hline
Italian American & yes (100)\\ \hline
Italian Democratic Socialist Party & yes (3), ``the It...'' found (9)\\ \hline
French  & yes (10538)\\ \hline
French language  & yes (336)\\ \hline
French spacing (English) & ``French Spacing'' found (3) \\ \hline
French fries & yes (107)\\ \hline
French Revolution & yes (149)\\ \hline
French and Indian War & yes (49)\\ \hline
French colonial empire & yes (10)\\ \hline
French Mandate for Syria and the Lebanon  & yes (2), ``the Fr..'' found (5) \\ \hline
French cuisine & yes (78) \\ \hline
French people & yes (105)\\ \hline
\end{tabular}
}
\end{center}
}
\caption{Query expansion suggestions for ``John'', ``Peter'', ``Blue'', ``Brown'', ``Italian'', and ``French'' from the English Wikipedia. Entries ``yes'' indicates that the same expansion was found automatically after indexing Wikipedia using above Step 3 ``kernel expansion''. The number of occurrences of a phrase is given in brackets.}\label{tableSmaller}
\end{table}

\begin{figure}
\begin{center}
\includegraphics[width=0.7\textwidth]{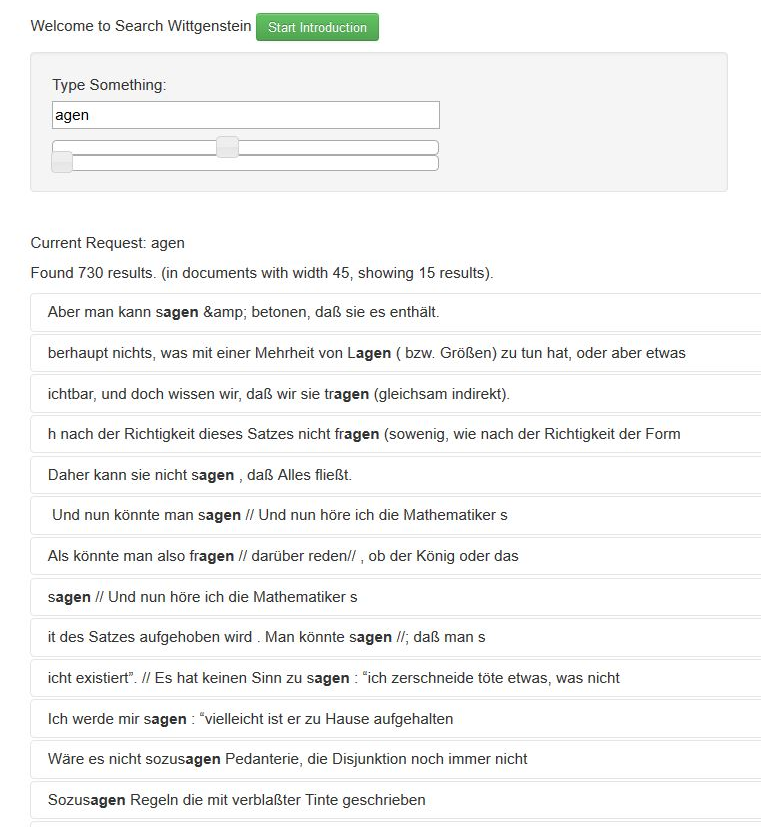}
\end{center}
\caption{Free context size answer browsing.}\label{Figanswerbrowsing}
\end{figure}

{\bf Free context size answer browsing.\ } Once the user has selected one of the expansion suggestions (``John McCain''), Wikipedia leads her to the Wikipedia page for the expansion (https://en.wikipedia.org/wiki/John$\_$McCain). The symmetric index technology can be used for another type of interaction, which we call ``free context size answer browsing''. We  present to the user a {\em two-sided} list of concordances that includes all occurrences of the expansion in the indexed corpus. Since in the index left and right contexts of arbitrary length are directly accessible, the {\em size of contexts shown can be selected and varied by the user}. In Figure~\ref{Figanswerbrowsing} we see a  user interface with this functionality embedded. 
In this search interface, after typing any string (here: ``agen''), the user may select the number of hits and the size of the concordances, which includes both left and right contexts for a hit. Using the lower button, the size of contexts presented can be changed in real-time. This kind of search interface is used at a project at CIS, LMU, where we collaborate with philosophers to find new and innovative ways of how to access corpora in Digital Humanities. See http://www.cis.uni-muenchen.de/dighum/research-group-co/index.html.


\section{Related work}\label{Sec5}

The discussion of related work is separated into two parts. We first consider contributions that are related when looking at the general characteristics of 
the approach presented above. Afterwards we take a more application oriented view. 

\subsubsection*{General characteristics of approach}

{\em Finding phrases.\ } When looking for phrases, we do not use linguistic knowledge and only consider structural properties of the corpus. Our approach is completely language independent. 
In many application fields, as a common standard approach phrases for simplicity are replaced by n-grams of words,~\cite{ON04,PK10}. Additional grammatical
features, e.g. part of speech~\cite{LY07}, syntactic~\cite{ZLC13}, or semantic markers~\cite{EWVN11} may be applied to find n-grams that exhibit phrase characteristics. For larger values of $n$, statistical uncertainty is growing. 
Deep learning approaches~\cite{CWB11,GJM13} try to resolve this issue by 
deriving regularities in smaller dimension spaces. However, the embedding of data in a smaller dimension space is achieved by extracting features from a fixed length context or from the entire sentence,~\cite{CWB11}, with no attempt to reflect the general structure of the corpus. Recent approaches in this area try to account for this drawback by more advanced network topology,~\cite{TSM15}.
Due to the fixed size of $n$-grams considered at one time, and since $n$-grams start at any point in the sentence, the subphrase structure of sentences is not analyzed.  
  


{\em Grammar development and induction.\ } 
When trying to analyze sentence structure, grammar-based parsing algorithms can be used \cite{tomita2013efficient}. 
The sentence and phrase decomposition strategies described in this paper can be considered as a preliminary step towards corpus based, automated, and language independent grammar induction. While our approach leads from corpus structure to grammatical rules, 
in parsing, fixed grammar rules impose a ``predefined'' structure on the corpus.   
For most languages, a full and correct syntactic analysis of arbitrary sentences in unrestricted texts is not possible. For searching phrases of a particular type, partial/local grammars  \cite{Gross97} can achieve high precision and recall. Until today, the development of (full, partial, or local) grammmars with high coverage and precision is an ambitious and difficult task that needs a high amount of human work. Furthermore, most methods we are aware of are language dependent. Language independent and automated construction steps are found in probabilistic context-free grammars \cite{sakakibara2010probabilistic} and learning regular expressions \cite{denis2004learning}. The latter problem is mainly of theoretical interest.

{\em Sequence analysis without linguistic knowledge.\ } 
The problem of detecting meaningful units in a text corpus can be considered also as a kind of segmentation problem. 
In~\cite{Liang:2009:OEU:1620754.1620843} and in~\cite{GGJ09} the authors consider the problem of segmenting sequences
of letters without blanks and punctuation into words. \cite{GGJ09} is focused on the segmentation of children's utterances transcriptions into words. 
The authors present a Bayesian approach combined with the Maximum A Posteriori Principle in order to detect the most probable boundaries of words 
within a sequence. In particular, they use Dirichlet distribution to model the number of different words contained in the corpus and a probability for an 
end of a word. 

The approach presented in~\cite{Liang:2009:OEU:1620754.1620843} is a generic one and the authors apply it also to Chinese word segmentation, 
Machine Translation alignment and others. The model defines only a probability for a sequence of letters to belong together. In this measure a key role 
is played by a fine-tuned parameter that controls the length of sequences. Other parameters of the model are trained via 
Expectation Maximisation Principle.

An extensive work has been also done on supervised word segmentation in Chinese. The methods range from statistical approaches based on CRFS~\cite{PFM04} to Recurrent Neural Networks~\cite{CQZH15} and others.

There are two main differences between the works cited above and the current paper. Firstly, the problem of segmentation requires that each letter is assigned to exactly one word. In contrast, we allow overlaps between these units which represents the possibility of sharing small pieces of information which however have high frequency and can provide us with reliable statistics. Secondly, it does not seem that previous approaches reflect the structure of the corpus, rather it is the statistics which aggregates the statistics obtained from the different entries of the corpus. In the presented approach we use this structure in a two-fold manner: (i) to constrain the number of available strings and thus define their possible boundaries in terms of other strings; and (ii) to find the best fit of these shorter strings within a particular corpus' entry. This considerations allow us to avoid the explicit modelling of phrase/word types and phrase/word length.   

{\em Index-based analysis of corpora.\ } 
In \cite{Goller2010} (see also \cite{Guenthner2005}), suffix arrays are used as a text index structure for mining local grammars. POS-info stored in a synchronized index structure enables a rich set of queries including textual and grammatical conditions. In many fields, suffix trees/arrays are used for various tasks, applications ranging from sequence analysis in bioinformatics \cite{gusfield1997algorithms} to text-reuse in digital humanities. The index structure used for our approach, symmetric directed acyclic word graphs (SDAWGs) \cite{Inenaga01}, has yet not found much attention. The power of this index relies on the fact that it gives access to left and right contexts of any length. This feature is used in many of the above algorithms. 

\subsubsection*{Applications}

We mentioned several possible applications of the methods developed. In all those fields there exists a large 
literature, only a rudimentary impression can be given. {\em Finding characteristic terms} of a document collection is studied, for example, in Information Retrieval. ``Global'' term weights are, e.g., Poisson overestimation, discrimination value, and inverse document frequency \cite{MRS07}. Specialized methods have been developed for the related problem of automated keyword extraction \cite{rose2010automatic}.
{\em Terminology extraction} is studied, for example, in \cite{patry2005corpus}. Methods for {\em automated query expansion} are discussed, e.g., in \cite{luke2012improving}. 
In corpus linguistics and lexicology, the problem of {\em finding syntagmatic and paradigmatic relations} 
is a central topic. \cite{SchuetzePedersen93,storjohann2010lexical} present contributions to finding lexical-semantic relations, collocation, word maps, and paradigmatics of a lexical unit. 
In all cases, the methods suggested above are unique by using information on the subphrase structure of sentences inferred from general structural properties of the corpus. For achieving optimal results it would be interesting to combine distinct approaches.

\section{Conclusion}\label{Sec6}

In this paper we described results of an ongoing research project 
where we analyze corpora in a completely language independent and unsupervised way without any prior linguistic knowledge. 
We introduced an algorithm for detecting function words and phrases. In many cases the phrases (combinations of phrases) obtained  are linguistic phrases or extensions with ``adjuncts'' (words connecting linguistic phrases). We then showed how phrases can be split into subphrases and how to determine content words. For Chinese, detection of phrases yields satisfactory results, as to word segmentation it is difficult to detect the best level of phrase decomposition. 

 
A weakness of the main algorithm for partitioning sentences into phrases presented in Sections~\ref{Sec2Sub1}-\ref{SubsectionQualitative} is that only strings can qualify as phrases that occur within distinct contexts in the corpus. All sequences with just one occurrence in the corpus cannot become phrases. To avoid this problem, and to come closer to phrases in a proper linguistic sense we need to {\em abstract} from 
plain symbol sequences in some way. The extensions of the main algorithm  presented in Section~\ref{SubsectionExtensions} represent one promising path for abstraction  that deserves further investigation. The detection of phrase patterns such as those in Section~\ref{SubsectionExtensions} and Section~\ref{Sec4} combined with the pure character-based structure underlying the approach may be also useful for recognising the morphological structure of the language and its dependence on the local phrase context. Some preliminary experiments in this direction showed that our index-structure provides structural indicators that can be used to extract reliable sets of morphemes and to segment words. We assume that unsupervised statistical methods for morphology induction, e.g.~\cite{Goldsmith01,Z08,creutz2005unsupervised}, can be combined with methods for mining the phrase structure of the corpus, providing another form of abstraction. This might provide a solid alternative to the deep neural network approach taken by~\cite{soricut2015unsupervised} where embeddings account for the local context of the words and their morphological properties. 


The applications discussed show that the unsupervised analysis of (sub)phrase structure and the two-directional symmetric index used for corpus representation offer many interesting options for finding new solutions to prominent problems in distinct fields.  

In this paper, when analyzing corpora we were radical in the sense that we do not use any kind of explicit linguistic knowledge/resource and no knowledge on the nature/functionality of a symbol (blank, hyphen,...). As a matter of fact, from a practical perspective it is interesting to give up this rigid principle. We intend to adapt the methods obtained to ``non-raw'' texts. For example, text annotated with POS info just represents another kind of interesting source data \cite{Goller2010} for ``language mining''. The active use of prior knowledge is one point of future work.

{\bf Acknowledgements.\ } We express our gratitude to: Max Hadersbeck for the sentence segmentation of the Accountant Corpus; Thomas M\"uller for providing us with sentence segmentation of the German and English dumps of Wikipedia from May 2014. Wenpeng Yin for the translations from Chinese to English. Stoyan Mihov, Petar Mitankin and Tinko Tinchev for comments and discussions on the ongoing research. Hinrich Sch\"utze for a discussion on function words.
The second co-author would like to explicitly acknowledge the technical lead of the first co-author: in the wake of many joint discussions, it was the first co-author who found and implemented the algorithms suggested in this paper. 
{\em This research was carried out within the ISALTeC Project, Grant Agreement 
625160 of Marie Curie Intra Fellowship within the 7th European Community Framework Programme.}

\bibliographystyle{alpha}
\bibliography{./bibliography_draft2}

\section*{Appendix A}

We present top segments of the ranked lists of terminological expressions obtained for some kernels using the method described in Section~\ref{SubSec4.1}. No filtering or cleansing steps were applied to improve the lists. 

\begin{example}
Terminological expressions obtained for kernel ``disease'':

{\tiny
disease/8418
disease./832
diseases/526
disease,/390
diseases./348
diseases,/158
this disease./153
coronary artery disease/85
infectious diseases/85
diseased/82
Alzheimer's disease./82
this disease/77
coronary artery disease./73
liver disease./68
cardiovascular disease./63
liver disease/53
Alzheimer's disease (AD)/52
Alzheimer's disease,/52
heart disease/52
Parkinson's disease./51
disease-specific/51
heart disease./50
disease)/47
Crohn's disease/46
disease-free survival/46
coronary heart disease./45
autoimmune diseases./45
metastatic disease./45
cardiovascular disease/43
Alzheimer's disease/42
diseases such/42
cardiovascular disease,/42
The disease/41
autoimmune diseases/41
vascular disease/40
cardiovascular diseases/40
Alzheimer's disease (AD)./39
coronary heart disease (CHD/39
coronary artery disease,/38
Parkinson's disease/37
cardiovascular diseases./37
disease progression/36
coronary heart disease/35
renal disease./34
metastatic disease/33
disease-related/33
liver disease,/31
infectious diseases./31
Parkinson's disease (PD)/31
Crohn's disease./30
disease-free/30
Parkinson's disease,/30
this disease,/29
course of the disease./29
inflammatory bowel disease/29
autoimmune disease/28
coronary heart disease,/28
vascular disease./28
diseases, including/28
inflammatory bowel disease./28
coronary artery disease (CAD)/28
Alzheimer's disease (AD/28
renal disease/27
disease-/27
Hodgkin's disease/26
cardiovascular disease (CVD/26
neurodegenerative diseases/26
Hodgkin's disease./25
periodontal disease./25
Crohn's disease,/25
neurodegenerative diseases./25
kidney disease./25
inflammatory diseases/25
disease)./24
obstructive pulmonary disease (COPD)./24
diseases, such/23
lung disease./23
Parkinson's disease (PD)./23
and disease./23
disease severity/23
lung disease/22
periodontal disease/22
liver diseases/22
inflammatory diseases./22
ischemic heart disease/22
disease, including/21
obstructive pulmonary disease (COPD/21
heart disease,/21
coeliac disease./20
ischaemic heart disease./20
disease;/20
autoimmune disease./20
coronary artery disease (CAD)./20
Graves' disease./20
cardiovascular diseases,/20
Parkinson's disease (PD/20
diseases in/19
congenital heart disease/19
various diseases/19
liver diseases./19
sickle cell disease./19
renal disease,/19
sexually transmitted diseases/19
advanced disease./19
end-stage renal disease (ESRD/19
respiratory diseases/19
diseases associated/19
and diseased/18
vascular diseases/18
inflammatory bowel disease,/18
graft-versus-host disease (GVHD/18
genetic diseases/18
kidney disease/18
kidney disease (CKD/18
chronic obstructive pulmonary disease (COPD/18
lung diseases/17
underlying disease./17
Paget's disease/17
allergic diseases/17
stable disease./17
cerebrovascular disease./17
infectious disease/17
disease-associated/17
end-stage renal disease./17
cerebrovascular disease,/17
pulmonary disease./17
infectious disease./17
malignant diseases/16
ischemic heart disease./16
disease, especially/16
malignant disease./16
disease, particularly/16
coronary disease./16
progressive disease/16
other diseases/16
kidney disease,/16
disease activity/16
course of the disease/15
vascular disease,/15
a disease/15
Hodgkin's disease,/15
cardiac disease/15
Graves' disease/15
congenital heart disease./15
periodontal diseases/15
chronic diseases/15
infectious diseases,/15
recurrent disease./15
the disease/15
severity of the disease./15
inflammatory bowel disease (IBD)/15
coronary heart disease (CHD)./15
various diseases./15
obstructive pulmonary disease./15
disease, and/15
recurrent disease/15
}
\end{example}

\begin{example}
Terminological expressions obtained for kernel ``syndrome'':

{\tiny syndrome/549
syndrome./287
syndrome,/173
syndromes/157
metabolic syndrome/100
syndromes./82
this syndrome/62
syndromes,/60
metabolic syndrome./47
nephrotic syndrome/44
this syndrome./33
acquired immunodeficiency syndrome (AIDS/33
metabolic syndrome,/31
Down syndrome/30
syndrome)/27
Down's syndrome/26
Sjögren's syndrome/26
The syndrome/22
irritable bowel syndrome/20
syndrome)./20
syndrome and/20
carpal tunnel syndrome/19
Cushing's syndrome/19
syndrome characterized/19
respiratory distress syndrome/17
a syndrome/16
acquired immune deficiency syndrome (AIDS)/16
Cushing's syndrome./16
compartment syndrome/16
This syndrome/15
irritable bowel syndrome (IBS)/14
long QT syndrome/13
Down syndrome./13
toxic shock syndrome/13
nephrotic syndrome./12
Cushing's syndrome,/12
Rett syndrome/12
myelodysplastic syndrome (MDS/12
meconium aspiration syndrome/11
clinical syndrome/11
syndrome associated/11
-syndrome/11
Budd-Chiari syndrome/11
Turner's syndrome/11
Down syndrome,/11
coronary syndromes/11
polycystic ovary syndrome (PCOS)/11
syndrome\&quot/11
Tourette syndrome/11
this syndrome,/10
Down's syndrome,/10
syndrome is/10
The metabolic syndrome/10
antiphospholipid syndrome/10
metabolic syndrome (MetS)/10
's syndrome/9
syndrome is presented./9
Wolff-Parkinson-White syndrome/9
Down syndrome (DS)/9
respiratory distress syndrome,/9
acute coronary syndrome/9
hepatopulmonary syndrome/9
acquired immunodeficiency syndrome/9
pain syndromes/8
Marfan's syndrome/8
adult respiratory distress syndrome/8
respiratory distress syndrome (RDS/8
nephrotic syndrome,/8
Reiter's syndrome/8
syndrome in/8
syndrome, including/8
Turner syndrome/8
Tourette's syndrome,/8
Marfan syndrome/8
myelodysplastic syndrome/8
coronary syndrome/8
Metabolic syndrome/8
pain syndrome/8
pain syndrome./8
Marfan syndrome./8
Lennox-Gastaut syndrome/8
sicca syndrome/8
Cockayne syndrome/8
Sheehan's syndrome/7
Goodpasture's syndrome/7
Sjögren's syndrome./7
respiratory distress syndrome./7
pain syndromes./7
acquired immunodeficiency syndrome (AIDS)./7
syndrome is characterized/7
Felty's syndrome/7
polycystic ovary syndrome/7
withdrawal syndrome/7
sleep apnoea syndrome/7
Terson syndrome/7
coronary syndromes./7
Williams syndrome/7
acquired immunodeficiency syndrome./7
coronary syndrome./7
fatigue syndrome (CFS/7
Rett syndrome./7
Behçet's syndrome/7
SVC syndrome/6
Bartter's syndrome/6
) syndrome/6
Zollinger-Ellison syndrome/6
Down's syndrome./6
syndrome occurred/6
adult respiratory distress syndrome (ARDS)/6
Tourette's syndrome/6
Wiskott-Aldrich syndrome/6
fragile X syndrome,/6
obstructive sleep apnea syndrome/6
sudden infant death syndrome (SIDS)./6
Guillain-Barré syndrome/6
compartment syndrome,/6
syndrome\&quot;./6
obstructive sleep apnea syndrome (OSAS)/6
syndromes such/6
respiratory distress syndrome (ARDS)./6
systemic inflammatory response syndrome/6
wasting syndrome/6
Lowe syndrome/6
metabolic syndrome (MS)/6
Brugada syndrome/6
Apert syndrome/6
Beckwith-Wiedemann syndrome/6
Ogilvie's syndrome/6
syndrome patients./6
fragile X-associated tremor/ataxia syndrome/6
Gilbert's syndrome/5
syndrome due/5
's syndrome,/5
deficiency syndrome/5
fat embolism syndrome/5
-like syndrome/5
syndrome (P/5
pain syndromes,/5
Reye's syndrome/5
adult respiratory distress syndrome./5
syndrome'/5
sick sinus syndrome/5
abstinence syndrome/5
Raynaud's syndrome/5
acquired immune deficiency syndrome (AIDS)./5
myelodysplastic syndromes/5
heart-asthenia syndrome/5
Sjögren's syndrome,/5
Marfan's syndrome./5
syndrome caused/5
syndromes, including/5
HELLP syndrome/5
The clinical syndrome/5
carpal tunnel syndrome./5
sudden infant death syndrome (SIDS)/5
short bowel syndrome./5
syndrome (/5
post-polio syndrome/5
Turner syndrome./5
antiphospholipid syndrome./5
Zollinger-Ellison syndrome (ZES)/5
fragile X syndrome./5
sudden infant death syndrome/5
Guillain-Barré syndrome,/5
syndrome (HUS)/5
syndrome are described./5
Angelman syndrome/5
Kallmann syndrome/5
systemic inflammatory response syndrome (SIRS)/5
acquired immune deficiency syndrome/5
syndrome occurs/5
Stevens-Johnson syndrome/5
Klinefelter's syndrome/5
Bartter syndrome/5
the metabolic syndrome/5
Panayiotopoulos syndrome/5
Noonan syndrome/5
polycystic ovary syndrome (PCOS)./5
polycystic ovary syndrome./5
Sweet's syndrome/5
metabolic syndrome and/5
acute coronary syndromes (ACS/5
syndrome was diagnosed/5
syndrome was diagnosed./5
syndrome-associated/5
WAGR syndrome/5
Churg-Strauss syndrome (CSS)/5
Sjogren's syndrome/5
post-Q-fever fatigue syndrome/5
coronary syndrome,/5
coronary syndromes (ACS/5
Hunter syndrome./5
Eisenmenger syndrome./5
syndrome, especially/5
scimitar syndrome/5
5q- syndrome/5
}
\end{example}

\begin{example}
Terminological expressions obtained for kernel ``inflammatory'':

{\tiny 
inflammatory/381
anti-inflammatory/117
inflammatory response/111
proinflammatory/63
proinflammatory cytokines/52
inflammatory response./46
antiinflammatory/41
pro-inflammatory/40
inflammatory reaction/37
inflammatory process/35
inflammatory cells/34
inflammatory mediators/33
inflammatory cytokines/33
inflammatory bowel/32
anti-inflammatory effects/31
inflammatory responses/31
pro-inflammatory cytokines/30
inflammatory bowel disease/29
inflammatory bowel disease./28
inflammatory cells./28
inflammatory markers/27
an inflammatory/26
inflammatory diseases/25
inflammatory lesions/24
proinflammatory cytokine/24
pro-inflammatory cytokine/23
inflammatory processes/23
chronic inflammatory/23
anti-inflammatory effect/22
inflammatory diseases./22
inflammatory processes./21
inflammatory responses./21
inflammatory changes/21
inflammatory response,/19
inflammatory process./19
inflammatory bowel disease,/18
inflammatory,/17
inflammatory reaction./17
inflammatory reactions/17
anti-inflammatory drugs/16
inflammatory disorders/15
inflammatory conditions./15
inflammatory bowel disease (IBD)/15
the inflammatory/15
inflammatory infiltrate/14
inflammatory cell infiltration/14
non-steroidal anti-inflammatory drugs (NSAIDs)/14
systemic inflammatory response/14
nonsteroidal anti-inflammatory drugs/13
inflammatory disease./13
nonsteroidal anti-inflammatory drugs (NSAIDs/13
inflammatory conditions/13
anti-inflammatory activity/12
inflammatory cytokine/12
anti-inflammatory properties/12
proinflammatory mediators/12
anti-inflammatory activity./12
inflammatory markers,/12
The inflammatory response/11
inflammatory mediators,/11
systemic inflammatory/11
inflammatory bowel disease (IBD)./11
inflammatory markers./11
inflammatory bowel disease (IBD/11
The inflammatory/10
inflammatory disease/10
proinflammatory cytokines./10
inflammatory process,/10
anti-inflammatory agents/9
inflammatory diseases,/9
inflammatory disorders./9
inflammatory bowel diseases/9
The anti-inflammatory activity/9
inflammatory cytokines./9
inflammatory cytokines,/9
pro- and anti-inflammatory/9
pelvic inflammatory/8
anti-inflammatory drugs./8
inflammatory demyelinating/8
noninflammatory/8
anti-inflammatory properties./8
anti-inflammatory effects./8
proinflammatory cytokines,/8
neuroinflammatory/8
inflammatory cell/7
inflammatory changes./7
inflammatory infiltrate./7
inflammatory mediator/7
The anti-inflammatory/7
inflammatory parameters/7
inflammatory diseases such/7
infiltration of inflammatory/7
nonsteroidal anti-inflammatory drug/7
inflammatory reactions./7
inflammatory cascade/7
anti-inflammatory cytokine IL-10/7
inflammatory disease,/7
inflammatory state/7
inflammatory activity/6
nonsteroidal anti-inflammatory drugs./6
inflammatory infiltrates/6
non-inflammatory/6
Anti-inflammatory/6
various inflammatory/6
anti-inflammatory cytokine/6
inflammatory pseudotumor/6
inflammatory pain/6
inflammatory events/6
inflammatory conditions,/6
inflammatory infiltration/6
inflammatory mediators such/6
systemic inflammatory response syndrome/6
inflammatory mediators./6
pelvic inflammatory disease,/6
non-steroidal anti-inflammatory drug (NSAID)/6
macrophage inflammatory protein-1alpha/6
of inflammatory/6
inflammatory states./6
pro-inflammatory cytokines,/6
pro-inflammatory cytokines./6
Pro-inflammatory cytokines/6
inflammatory responses,/6
non-steroidal anti-inflammatory drugs/6
macrophage inflammatory/6
inflammatory changes,/6
Proinflammatory cytokines/6
anti-inflammatory agents./5
fibro-inflammatory/5
pelvic inflammatory disease (PID)/5
inflammatory pain./5
nonsteroidal antiinflammatory/5
inflammatory reactions,/5
necroinflammatory/5
inflammatory and neoplastic/5
inflammatory skin/5
inflammatory lung/5
An inflammatory/5
proinflammatory mediators,/5
non-specific inflammatory/5
systemic inflammatory response syndrome (SIRS)/5
inflammatory reaction,/5
proinflammatory cytokine,/5
active inflammatory bowel/5
nonsteroidal anti-inflammatory drugs,/5
inflammatory disorder/5
inflammatory stimuli/5
postinflammatory/5
non-steroidal anti-inflammatory drugs,/5
production of proinflammatory/5
inflammatory component./5
systemic inflammatory response./5
several inflammatory/5
inflammatory status/5
inflammatory cell infiltration./5
inflammatory signaling/5
anti-inflammatory activities./5
inflammatory stimuli./5
non-steroidal anti-inflammatory drug/5
nonsteroidal antiinflammatory drugs (NSAIDs)/5
release of inflammatory mediators/5
}
\end{example}

\end{document}